\documentclass[lettersize,journal]{IEEEtran}
\usepackage{amsmath,amsfonts}
\usepackage{algorithmic}
\usepackage{algorithm}
\usepackage{array,soul}
\usepackage[caption=false,font=normalsize,labelfont=sf,textfont=sf]{subfig}
\usepackage{textcomp}
\usepackage{stfloats}
\usepackage{url}
\usepackage{verbatim}
\usepackage{graphicx}
\usepackage{cite}
\usepackage{booktabs}
\usepackage{multirow} 
\usepackage{xcolor}
\usepackage[colorlinks,linkcolor=blue]{hyperref}
\newcommand{\model}{OmniScene}
\newcommand{\nuScenes}{nuScenes}

\def\BibTeX{{\rm B\kern-.05em{\sc i\kern-.025em b}\kern-.08em
    T\kern-.1667em\lower.7ex\hbox{E}\kern-.125emX}}
\usepackage{balance}

\begin{document}
% \title{{\model}: Enhancing End-to-End Autonomous Driving with Multimodal Driver Attention Fusion}

% \title{DFA-Driving: Attention-Augmented 3D Scene Understanding for End-to-End Driving via Human-aligned Multimodal Fusion}

\title{{\model}: Attention-Augmented Multimodal 4D Scene Understanding for Autonomous Driving}

\author{
Pei Liu, Hongliang Lu, Haichao Liu, Haipeng Liu, Xin Liu, Ruoyu Yao, \\ Shengbo~Eben~Li, \textit{Senior Member, IEEE,} Jun Ma, \textit{Senior Member, IEEE}
        % <-this % stops a space
        \thanks{Pei Liu and Hongliang Lu contributed equally to this work.}
\thanks{Pei Liu, Hongliang Lu, Haichao Liu, Xin Liu, Ruoyu Yao, and Jun Ma are with The Hong Kong University of Science and Technology, China (e-mail: \{pliu061, hlu592, hliu369, xliu969, ryao092\}@connect.hkust-gz.edu.cn;  jun.ma@ust.hk).}% <-this % stops a space
\thanks{Haipeng Liu is with Li Auto Inc., Shanghai 201800, China (e-mail: liuhaipeng@lixiang.com).}
\thanks{Shengbo Eben Li is with the School of Vehicle and Mobility, Tsinghua University, Beijing 100084, China. (e-mail: lishbo@tsinghua.edu.cn).} 
% \thanks{Jun Ma is with The Hong Kong University of Science and Technology (Guangzhou), Guangzhou 511453, China, and also with The Hong Kong University of Science and Technology, Hong Kong SAR, China (e-mail: jun.ma@ust.hk).}

% \thanks{Corresponding author: Jun Ma.}
}

% \markboth{Journal of \LaTeX\ Class Files,~Vol.~18, No.~9, September~2020}%
% {How to Use the IEEEtran \LaTeX \ Templates}

\maketitle

\begin{abstract}
    Human vision is capable of transforming two-dimensional observations into an egocentric three-dimensional scene understanding, which underpins the ability to translate complex scenes and exhibit adaptive behaviors. This capability, however, is still lacking in current autonomous driving systems, where mainstream approaches largely rely on depth-based 3D reconstruction rather than true scene understanding. To address this limitation, we propose a novel human-like framework called {\model}. First, we introduce the OmniScene Vision-Language Model (OmniVLM), a vision-language framework that integrates multi-view and temporal perception for holistic 4D scene understanding. Then, harnessing a teacher-student OmniVLM architecture and knowledge distillation, we embed textual representations into 3D instance features for semantic supervision, enriching feature learning, and explicitly capturing human-like attentional semantics. These feature representations are further aligned with human driving behaviors, forming a more human-like perception–understanding–action architecture.
    In addition, we propose a Hierarchical Fusion Strategy (HFS) to address imbalances in modality contributions during multimodal integration. Our approach adaptively calibrates the relative significance of geometric and semantic features at multiple abstraction levels, enabling the synergistic use of complementary cues from visual and textual modalities. This learnable dynamic fusion enables a more nuanced and effective exploitation of heterogeneous information. %, ultimately enhancing the model’s capacity for comprehensive scene understanding.
We evaluate {\model} comprehensively on the nuScenes dataset, benchmarking it against over ten state-of-the-art models across various tasks. Our approach consistently achieves superior results, establishing new benchmarks in perception, prediction, planning, and visual question answering. Notably, {\model} yields a remarkable 21.40\% improvement in visual question answering (VQA) performance, highlighting its robust multimodal reasoning capabilities.
Project Link: \href{https://github.com/ocean-luna/OmniScene}{https://github.com/ocean-luna/OmniScene}.
\end{abstract}


\section{Introduction}
\label{sec:intro}
Recent years have seen substantial advancements in autonomous driving, marked by progress across core domains including perception \cite{philion2020lift, li2022bevformer, liao2022maptr, liu2025revisiting}, motion prediction \cite{chai2019multipath, gu2023vip3d, jiang2022perceive}, and planning \cite{toromanoff2020end, prakash2021multi}. These breakthroughs have collectively reinforced the foundation for more precise and safer driving performance \cite{wang2022detr3d, jiang2023vad, hu2023planning}. Within this context, end-to-end (E2E) autonomous driving has gained prominence as an innovative paradigm. By harnessing extensive datasets, E2E approaches learn to map raw sensor inputs directly to predicted planning trajectories, thereby removing reliance on manual intermediate processing stages and enhancing both adaptability and scalability \cite{chen2024end}
However, classic E2E autonomous driving systems often generate future planned trajectories or low‑level control commands without effectively integrating perception and scene understanding. This lack of integration limits their ability to incorporate essential contextual information, such as traffic dynamics and navigation constraints, which are critical for robust autonomous driving. Such limitations become particularly evident in complex and ambiguous scenarios, where independent perception or simplistic prediction is insufficient for scene understanding, such as dealing with nuanced traffic interactions or adhering to traffic rules.
In contrast, human vision continuously transforms perceptual inputs into scene understanding, adapting its attention to the evolving driving contexts, such as traffic signals, pedestrian activities, and lane markers \cite{botvinick2008hierarchical, koechlin2003architecture, badre2008cognitive}. This attention-aware scene understanding plays a pivotal role in shaping humans’ superior driving capabilities. Therefore, a unified approach to enable human-like scene understanding is essential for intelligent and safe planning in autonomous driving systems.

\begin{figure}
    \centering
    \includegraphics[width=1.0\linewidth]{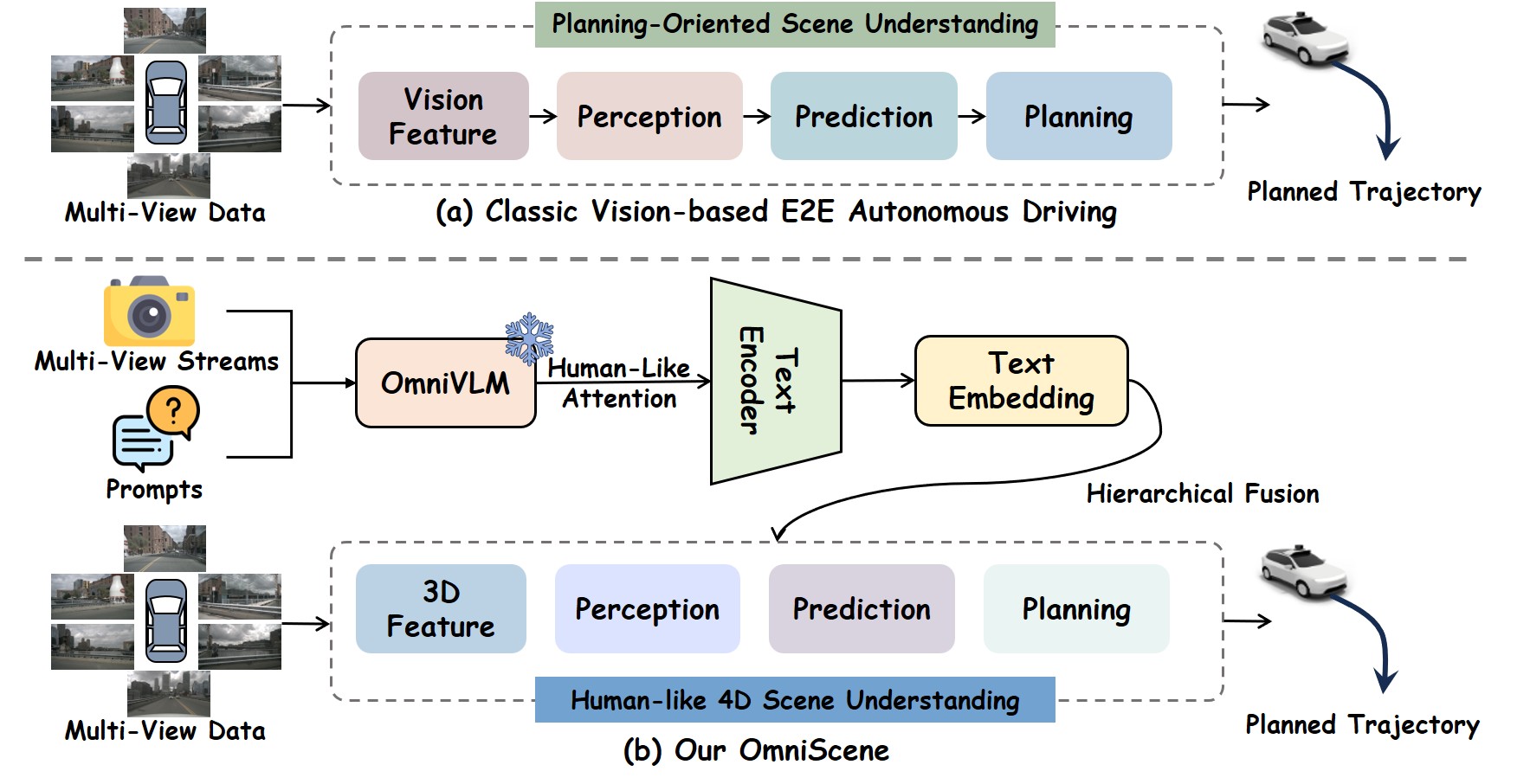}
    \caption{{\model} augments the end-to-end driving model with semantic textual descriptions during training. These descriptions extract human-like attention from VLMs to encourage the model to learn richer attentional semantics.}
    \label{fig:fig1}
\end{figure}

Recent advances in attention-aware planning have sought to enhance E2E autonomous driving by incorporating mechanisms such as self-attention, spatial attention, and local feature extraction modules \cite{chitta2022transfuser, li2024ego}. Despite these efforts, current methods often rely on low-level features or static heuristics, lacking explicit human-like attention modeling and failing to adapt in complex, evolving environments. More importantly, even with the emergence of Vision-Language Models (VLMs) \cite{liu2023visual, wang2025cogvlm, chen2024internvl, achiam2023gpt} that offer strong semantic abstraction, multimodal fusion remains superficial: visual and textual modalities are typically processed independently or in succession, rather than being deeply integrated.
This limitation results in underutilization of complementary information, as high-level semantics, attentional reasoning, and geometric context are not adequately combined to inform planning. Effective scene understanding for autonomous driving thus calls for human-aligned multimodal fusion strategies that can jointly aggregate 3D, visual, and semantic features, enabling more human-like context awareness and prioritization within dynamic driving scenarios.

Motivated by these challenges, we propose {\model} (depicted in Fig. \ref{fig:fig1}), an innovative framework designed to advance autonomous driving systems through human-like scene comprehension. This approach is designed to address three core issues:

\textbf{How to enable 4D scene understanding?}
Attaining robust 4D scene understanding necessitates the synthesis of perceptual and conceptual representations, bridging the gap between the raw geometric structure extracted from visual sensors and the high-level semantic interpretation characteristic of human cognition. While 3D geometric features capture the spatial configuration and dynamic relationships within the scene, textual semantic features encode context, intent, and abstract reasoning about environmental elements \cite{ji2024jm3d}. This dual-faceted integration reflects how humans interpret visual stimuli, in which sensory information is continually mediated by cognitive inference to support driving in complex, dynamic environments.
In our approach, multi-view 3D geometric features derived from sensor data reconstruct the spatial layout and motion states of objects, providing a precise foundation for tasks such as localization, obstacle avoidance, and motion planning. Complementing this, semantic features generated by a large VLM offer a higher-order understanding of attentional cues, navigation goals, and potential risks, supplying the contextual awareness needed for human-like judgment.
The fusion of these complementary modalities yields a unified representation that enables the autonomous system not only to “see” its environment with geometric accuracy but also to “understand” the scene in a manner analogous to human reasoning. This paradigm enhances both the interpretability and robustness of scene understanding, empowering autonomous driving systems to make informed and reliable decisions in complex traffic scenarios.

\textbf{How to enable human-like attention in scene understanding?}
Achieving human-like attention in scene understanding involves more than passive perception; it requires selective prioritization and contextual interpretation of visual cues, much like how expert drivers allocate cognitive resources in complex environments. In our framework, this capability is realized through the Omni-Scene Vision-Language Model (OmniVLM), which is specifically designed to process multi-view and multi-frame visual inputs for comprehensive scene perception and attentional reasoning.
Powered by advanced semantic reasoning abilities and large-scale multimodal knowledge, OmniVLM can generate attentional descriptions and decision rationales directly from parsed sensory inputs and environmental annotations that span different viewpoints and temporal frames. These outputs capture not only explicit scene elements but also latent dependencies and task-relevant priorities, closely resembling the nuanced attentional maps formed during human observation and inference.
To enable efficient deployment, we design a teacher–student OmniVLM architecture. Specifically, the original large-scale OmniVLM serves as the teacher model, transferring its attentional knowledge into a lightweight student model, such as spatial attention distributions and corresponding semantic rationales. Through knowledge distillation, the student OmniVLM learns to selectively focus on critical regions such as crosswalks, traffic signals, and nearby pedestrians, while suppressing irrelevant background information, much like the attentional mechanisms of human perception.
As a result, OmniVLM achieves robust and interpretable scene understanding with human-like attentional behavior, grounded in both geometric realism and semantic abstraction. This enables the development of an attention-aware driving agent capable of nuanced, context-sensitive reasoning and adaptive driving in dynamic and safety-critical scenarios.

\textbf{How to enable multimodal learning for E2E autonomous driving?} 
While general 3D scene understanding focuses on the reconstruction and interpretation of geometric structures and object relationships in space, autonomous driving requires more: accurate perception of spatial layout must be closely intertwined with semantic interpretation and context-aware reasoning. In real-world driving environments, the agent is tasked not only with modeling the positions and motions of diverse dynamic and static entities, but also with understanding their semantic significance and anticipating their evolution over time.
To meet these requirements, we develop a Hierarchical Fusion Strategy (HFS) that extends beyond conventional geometric analysis. Our approach integrates object-centric 3D instance representations with multi-view visual inputs and semantic attention derived from textual cues, anchored by explicit modeling of temporal dependencies. This multi-layered framework allows for a unified representation that captures both fine-grained spatial structures and high-level, temporal semantic priorities. By aligning the strengths of 4D reasoning with the adaptive interpretation of context and intent, our method advances the frontier of scene understanding in autonomous driving.
We test {\model} on {\nuScenes} \cite{caesar2020nuscenes}, a widely-used benchmark dataset for autonomous driving evaluation. Compared with over ten state-of-the-art models, our approach achieves significant improvements, demonstrating its effectiveness in enhancing perception, planning, and overall driving performance. %The key contributions of this work are summarized as follows:

% \begin{itemize}
% \item Augmented 3D Scene Understanding: We propose {\model}, which integrates perceptual and conceptual representations to achieve robust 3D scene understanding. By combining multi-view geometric features with high-level semantic information, our framework enables the autonomous system to accurately construct spatial layouts and interpret contextual cues, reminiscent of human cognitive processes in complex driving environments.

% \item Human-like Attention Mechanism: We propose an OmniVLM-based teacher-student framework to equip autonomous systems with human-like attention. The large-scale teacher OmniVLM generates rich attentional maps and decision cues from multi-view and multi-frame inputs. Through knowledge distillation, the lightweight OmniVLM student learns to selectively focus on critical visual cues, such as traffic signals and pedestrians, while filtering irrelevant information, thereby achieving efficient and human-like scene understanding and decision making.

% \item Multimodal Learning for End-to-End Driving: Our HFS effectively integrates 3D instance representations with visual and textual modalities, accounting for temporal dependencies. This approach ensures that the autonomous driving model not only understands the spatial organization of the environment but also grasps the semantic significance of dynamic entities. As a result, {\model} achieves superior adaptability and robustness in complex, real-world driving scenarios.

% \end{itemize}

The structure of this paper is as follows. Section \ref{sec:related} reviews
the relevant literature. Section \ref{sec:method} introduces the methodology. Section \ref{sec:experiments} details the experimental settings. Section \ref{result} presents the evaluation results. Finally, Section \ref{conclusion} concludes with a summary of the research.

\section{Related Work}
\label{sec:related}

\subsection{Multimodal Information Fusion Mechanism}
In recent years, attention-based fusion mechanisms and learnable fusion strategies have emerged as dominant paradigms for multi-modal information fusion, addressing the challenges of modality heterogeneity and imbalance. These approaches have demonstrated remarkable success in capturing cross-modal interactions and dynamically adapting to the relevance of each modality, making them particularly suitable for complex tasks such as autonomous driving and robotics.

Attention-based fusion mechanisms leverage the power of attention to model dependencies between modalities, enabling the model to focus on the most informative features. Transformer-based architectures \cite{vaswani2017attention, han2022survey, xu2023multimodal} have become a cornerstone of this approach, utilizing self-attention and cross-attention mechanisms to fuse features from different modalities. For instance, TransFuser \cite{chitta2022transfuser} employs transformers to integrate visual and LiDAR features, achieving state-of-the-art performance in 3D object detection and scene understanding. Similarly, cross-modal attention networks \cite{xu2020cross} use attention to weigh the importance of visual and textual features, enhancing tasks such as image-text matching and visual question answering. These methods excel at capturing long-range dependencies and complex interactions between modalities. However, they often require significant computational resources, limiting their applicability in real-time systems.

On the other hand, learnable fusion mechanisms have gained traction for their ability to dynamically adjust the contribution of each modality based on task-specific requirements. These methods introduce learnable parameters, such as weights or coefficients, to adaptively fuse features during training. For example, Modality-Aware Fusion \cite{liu2024cross} proposes learnable coefficients to balance the importance of visual and LiDAR features, improving robustness in autonomous driving tasks. Another notable approach is Dynamic Fusion Networks \cite{xue2023dynamic}, which uses gating mechanisms to selectively combine modalities based on their relevance to the current context. These strategies are particularly effective in handling modality imbalance, where one modality may dominate due to its inherent information richness or task-specific importance. By dynamically adjusting the fusion process, learnable mechanisms ensure that all modalities contribute meaningfully to the final output, enhancing both performance and interpretability.

\subsection{End-to-End Autonomous Driving} 
End-to-end autonomous driving systems have demonstrated significant improvements in overall performance by jointly training all modules under a unified objective, thereby minimizing information loss across the pipeline. In recent years, unified frameworks such as ST-P3 \cite{hu2022st} and UniAD \cite{hu2023planning} have pioneered vision-based E2E systems that seamlessly integrate perception, prediction, and planning modules, achieving state-of-the-art results in complex driving scenarios. Building on these advancements, subsequent research, such as VAD \cite{jiang2023vad} and VADv2 \cite{chen2024vadv2}, introduces vectorized encoding methods to enhance the efficiency and scalability of scene representation, enabling more robust handling of dynamic environments.

More recently, methods such as Ego-MLP \cite{zhai2023rethinking}, BEV-Planner \cite{li2024ego}, and PARA-Drive \cite{weng2024drive} have explored novel design spaces within modular stacks, focusing on self-state modeling and innovative architectural designs to further enhance driving performance. These approaches have pushed the boundaries of E2E systems by incorporating richer representations of the ego vehicle’s state and its interactions with the environment.

In this work, we build upon vision-based E2E autonomous driving by integrating human-like attentional text information. By leveraging natural language descriptions of critical driving cues, such as pedestrian crossing ahead or a red traffic light, we enable the model to explicitly capture and prioritize regions of interest that align with human-like attention. This enhancement not only improves the interpretability of the system but also ensures that the model’s decisions are more closely aligned with human-like reasoning, particularly in safety-critical scenarios.

\subsection{Vision Language Models in Autonomous Driving}
Despite the remarkable progress of VLMs in broad tasks, their application to autonomous driving raises several unique challenges. These challenges stem from the necessity to infuse models with driving-specific knowledge, accurately interpret complex traffic scenarios, and ensure that outputs align with the real-time safety and reasoning requirements of autonomous systems.

A primary challenge is the effective incorporation of driving-specific text prompts that convey the unique semantics and attentional cues within driving environments. Unlike general vision-language tasks, autonomous driving requires the model to understand nuanced instructions, such as “yield to pedestrians at crosswalks” or “brake for red lights ahead,” and to dynamically adapt its reasoning to safety-critical cues. Existing VLM-based systems often employ generic prompts or rely on large-scale vision-language pre-training, which may not sufficiently capture context-specific information crucial for safe driving decision-making.

% Moreover, integrating VLMs into end-to-end autonomous driving pipelines introduces further difficulties. Methods such as Drive-with-LLMs \cite{chen2024driving} and DriveGPT4 \cite{xu2024drivegpt4} have demonstrated the feasibility of leveraging VLMs for trajectory prediction and planning. However, these approaches often depend on ground-truth perception data or domain-specific finetuning, limiting their generalization to diverse real-world scenarios. Other works, such as ELM \cite{zhou2024embodied} and DriveVLM \cite{tian2024drivevlm}, highlight the importance of large-scale, cross-domain pre-training, but challenges remain in aligning model outputs with human-like decision processes and interpretability.

Moreover, integrating VLMs into end-to-end autonomous driving pipelines introduces further difficulties. Methods such as Drive-with-LLMs \cite{chen2024driving} and DriveGPT4 \cite{xu2024drivegpt4} have demonstrated the feasibility of leveraging VLMs for trajectory prediction and planning. However, these approaches often depend on ground-truth perception data or domain-specific finetuning, limiting their generalization to diverse real-world scenarios. Other works, such as ELM \cite{zhou2024embodied} and DriveVLM \cite{tian2024drivevlm}, highlight the importance of large-scale, cross-domain pre-training, but challenges remain in aligning model outputs with human-like decision processes and interpretability. Similarly, VLM‑E2E \cite{liu2025vlm} explores multimodal driver attention fusion within the BEV space, yet BEV-based integration can lose fine-grained 3D spatial context and weaken semantic–geometric alignment.

Another critical issue is the lack of high-quality, driving-centric vision-language datasets tailored to the complexities of urban and highway environments. While recent efforts \cite{sima2024drivelm, qian2024nuscenes, wu2023referring, kim2018textual} have begun to address this gap, further work is needed to capture rare, long-tail, or safety-critical scenarios that are essential for robust model behavior.
In summary, while VLMs offer promising capabilities for autonomous driving, advancing their application requires targeted solutions to address domain-specific semantics, data scarcity, real-time interpretability, and integration challenges. Our work aims to bridge these gaps by designing driving-attentional prompts and developing novel approaches for end-to-end vision-language reasoning in safety-critical driving scenarios.
\begin{figure*}
    \centering
    \includegraphics[width=1.0\linewidth]{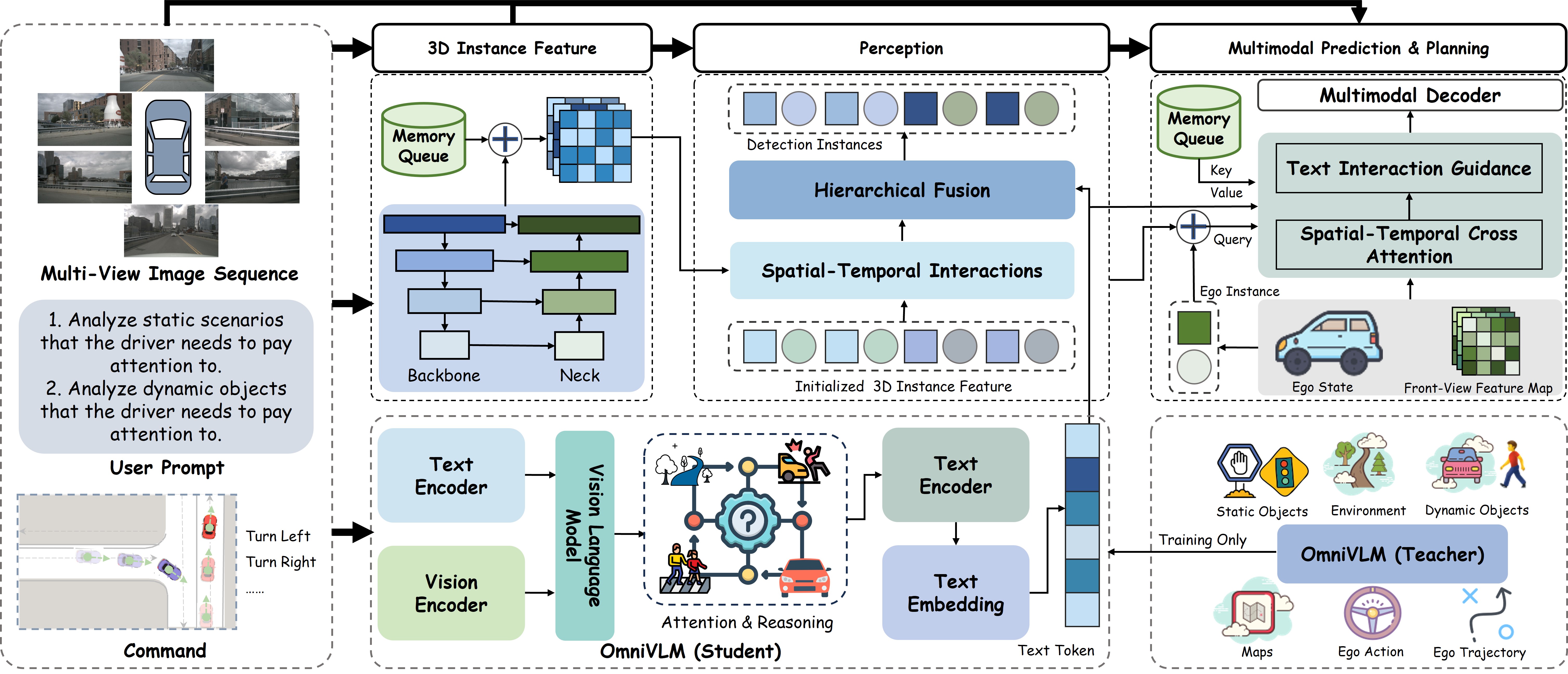}
    \caption{We present {\model}, a driver attention enhanced end-to-end vision-based framework. {\model} consists of three modules: Vision-based End-to-end Model, Hierarchical Fusion Module, and Teacher–Student OmniVLM Architecture.}
    \label{fig:architect}
\end{figure*}

\section{Methodology}
\label{sec:method}
In this section, we present a comprehensive overview of {\model}, as illustrated in Fig. \ref{fig:architect}. The input to the system consists of multi-view image streams, operational commands, and user prompts. These multimodal inputs are first processed by the student OmniVLM module, which generates concise textual annotations describing the observed scene. Simultaneously, the multi-view images are passed through a visual encoding layer to extract visual features. The generated textual annotations are then input to the HFS Module, where they are transformed into textual feature representations using a pre-trained CLIP model. Subsequently, the 3D instance features, visual features, and textual features are fused to provide comprehensive representations, supporting downstream tasks such as perception, prediction, and planning.

% Section \ref{sec:vlm} provides a detailed description of the teacher–student dual-OmniVLM architecture. Section \ref{sec:e2e} focuses on the design of the HFS Module, while Section \ref{sec:text-e2e} elaborates on the vision-based E2E model.

\subsection{Preliminary}
From an information-theoretic standpoint, multimodal aggregation can be formally characterized by analyzing how much complementary knowledge from vision and language is captured within the final 3D instance representation. Let $\mathcal{B}, \mathcal{I}$, and $\mathcal{T}$ denote the random variables for 3D instance, vision, and text modalities, respectively. The effectiveness of aggregation is measured by the total mutual information $I(\mathcal{B}; \mathcal{I}, \mathcal{T})$ between the aggregated 3D representation and the combined set of vision and text features, reflecting the model’s ability to integrate and preserve cross-modal semantic information.

A principled objective is to increase the mutual information $I(\mathcal{B}; \mathcal{I}, \mathcal{T})$ between the 3D instance and the collection of vision and text features, which can be decomposed as:
\begin{equation}
    I(\mathcal{B}; \mathcal{I}, \mathcal{T}) = I(\mathcal{B}; \mathcal{I}) + I(\mathcal{B}; \mathcal{T} | \mathcal{I}),
    \label{eq:1}
\end{equation}
where $I(\mathcal{B}; \mathcal{I})$ measures the shared information between the 3D instance and the vision, and $I(\mathcal{B}; \mathcal{T} | \mathcal{I})$ represents additional information provided by the text, conditional on the vision. In ideal aggregation, both terms are increased, indicating effective fusion.

During embedding learning, in addition to contrastive alignment, we consider minimizing the conditional entropy $H(\mathcal{B} | \mathcal{I}, \mathcal{T})$, which reflects uncertainty in the 3D instance given both vision and text modalities. A lower conditional entropy corresponds to reduced uncertainty in the fused 3D representation, thereby indicating more effective aggregation:
\begin{equation}
    H(\mathcal{B}| \mathcal{I}, \mathcal{T}) = -\mathbb{E}_{p(\mathcal{B}, \mathcal{I}, \mathcal{T})}[\log p(\mathcal{B}| \mathcal{I}, \mathcal{T})].
    \label{eq:2}
\end{equation}
It is pertinent to note that minimizing this entropy leads to representations where the 3D instance is highly predictable based on visual and textual cues.

Furthermore, to avoid redundancy and ensure that each modality contributes unique information, the interaction information may be considered:
\begin{equation}
    I(\mathcal{B}; \mathcal{I}; \mathcal{T}) = I(\mathcal{B}; \mathcal{I}) - I(\mathcal{B}; \mathcal{I}| \mathcal{T}).
    \label{eq:3}
\end{equation}
This term captures the net synergy between modalities in relation to the 3D instance. A positive value indicates that combined modalities provide more integrated information about the instance than either one alone.

\subsubsection{Maximizing Mutual Information}

Enhancing mutual information $I(\mathcal{B}; \mathcal{I}, \mathcal{T})$ is achieved through strategies that align multimodal features with 3D representations. Focal Loss for classification emphasizes rare or critical instances by disproportionately penalizing misclassification errors, ensuring the alignment of 3D features with visual features and textual cues. This also strengthens semantic correspondence, which effectively increases the mutual information components $I(\mathcal{B}; \mathcal{I})$ and $I(\mathcal{B}; \mathcal{T} | \mathcal{I})$.

Additionally, text conditional aggregation plays a pivotal role by embedding textual semantics into the learning process. This mechanism reduces redundancy between modalities and enhances interaction information $I(\mathcal{B}; \mathcal{I}; \mathcal{T})$, ensuring a synergistic integration that enriches 3D representations.

% \subsubsection{How to reduce conditional entropy $I(\mathcal{B}|\mathcal{I}, \mathcal{T})$ ?}
% The conditional entropy $I(\mathcal{B}|\mathcal{I}, \mathcal{T})$ measures uncertainty in 3D instance representations when given visual and textual cues (Equation \ref{eq:2}). Minimizing this entropy is critical for reducing ambiguity in 3D perception (e.g., precise localization of objects) and ensuring reliable downstream planning. Practical regression losses directly act as proxies for this objective:

% \begin{itemize}
%     \item L1 Loss for Regression: For 3D detection regression: This loss minimizes the error between predicted 3D bounding box parameters ($x_i$, $b_i$ for position and size) and ground truth, conditioned on multi-view visual features ($\mathcal{I}$) and textual cues ($\mathcal{T}$, e.g., "parked car 5m ahead"). By reducing regression error, L1 Loss ensures that 3D instance features ($\mathcal{B}$) are highly predictable from $\mathcal{I}$ and $\mathcal{T}$, directly lowering $H(\mathcal{B}|\mathcal{I}, \mathcal{T})$, as uncertainty about 3D object geometry decreases.

%     \item For motion prediction regression: This L1 loss minimizes displacement error between predicted trajectories and ground truth, conditioned on temporal visual features ($\mathcal{I}$, e.g., historical multi-frame images) and textual cues ($\mathcal{T}$, e.g., "slow down for turning vehicle"). By improving trajectory accuracy, the loss ensures that 3D instance dynamics (a key component of $\mathcal{B}$) are predictable from multimodal cues—again reducing conditional entropy.
% \end{itemize}

\subsubsection{Minimizing Conditional Entropy}

Reducing $H(\mathcal{B}|\mathcal{I}, \mathcal{T})$ is indispensable for achieving precise 3D predictions by lowering ambiguity in the fused representation. Regression objectives such as L1 Loss directly minimize prediction errors, applying to 3D bounding boxes and trajectory predictions conditioned on multimodal information. This reduction in geometric and dynamic uncertainty lowers entropy, which yields more reliable 3D instances.

Specifically, trajectory prediction losses minimize displacement errors by leveraging temporal visual cues and textual instructions (e.g., ``turning vehicle ahead"), thereby decreasing uncertainty in motion dynamics and further refining the accuracy of 3D representations.

\subsubsection{Unified Optimization for Cross-Modal Objectives}
The overall training objective (\ref{eq:train}) integrates classification (e.g., Focal Loss), regression (e.g., L1 Loss), and auxiliary objectives to simultaneously maximize $I(\mathcal{B}; \mathcal{I}, \mathcal{T})$ and minimize $H(\mathcal{B}|\mathcal{I}, \mathcal{T})$. Auxiliary objectives, such as depth alignment losses, promote consistent cross-modal information integration while avoiding modality-specific bias. This unified framework ensures strong semantic alignment, minimal redundancy, and reduced uncertainty in 3D representations, enabling robust and interpretable multimodal learning for downstream tasks.

% The overall training objective \ref{eq:train} integrates all practical losses to optimize the information-theoretic goals jointly:
% Classification losses (Focal Loss) maximize $I(\mathcal{B}; \mathcal{I}, \mathcal{T})$ by aligning 3D instance semantics with vision and text.  Regression losses (L1 Loss) minimize $H(\mathcal{B}|\mathcal{I}, \mathcal{T})$
%  by reducing uncertainty in 3D geometry and dynamics.
% Auxiliary losses (e.g., depth alignment loss) reinforce cross-modal consistency, ensuring that the model does not prioritize one modality over the others, thus maintaining a balance between maximizing mutual information and minimizing conditional entropy.

\subsection{Teacher–Student OmniVLM Architecture}
\label{sec:vlm}
\begin{figure*}
    \centering
    \includegraphics[width=1.0\linewidth]{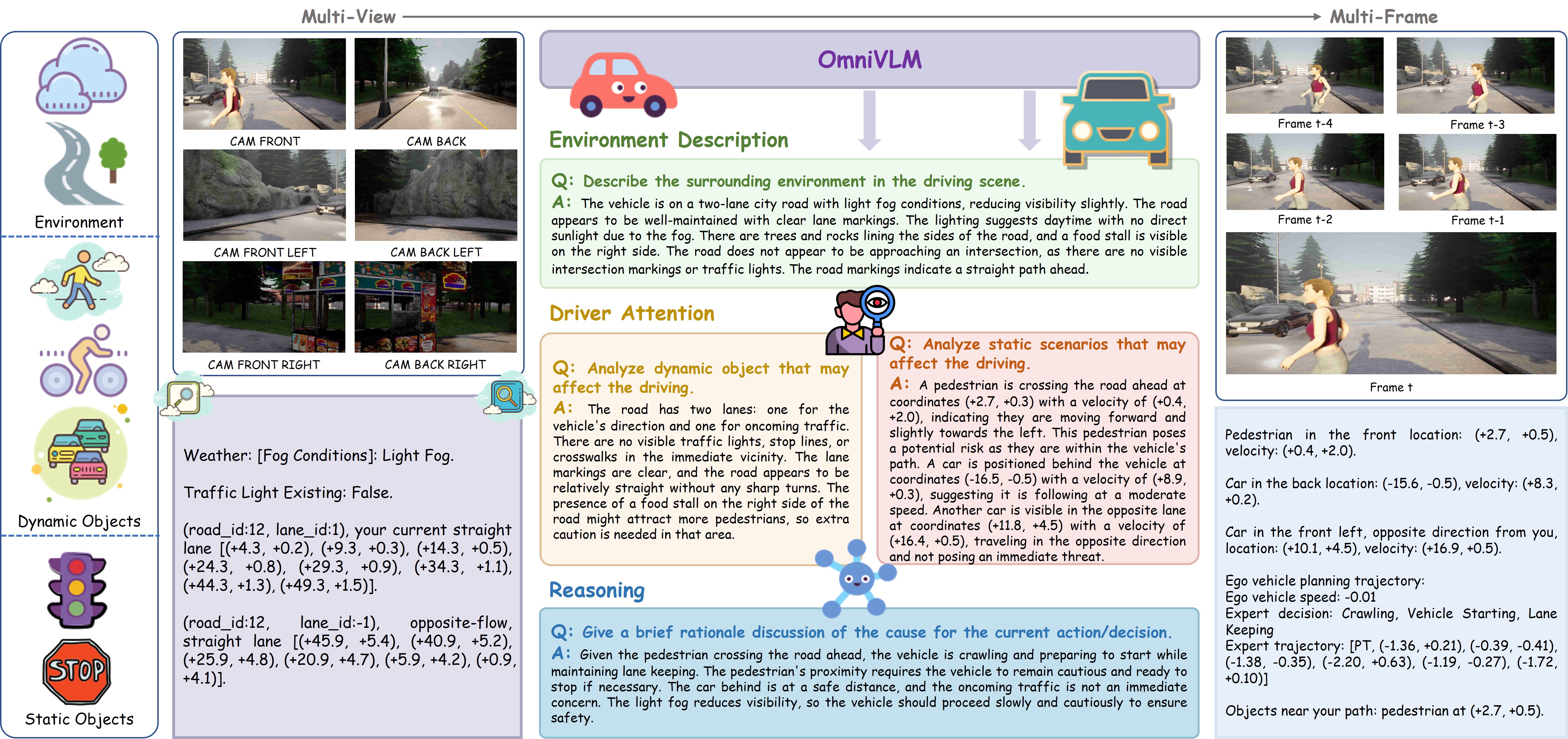}
    \caption{Overview of the teacher OmniVLM pipeline. Knowledge mining extracts ground truth, maneuvering signals, and domain rules from nuScenes data. These cues are then used for automated text generation, forming paired visual-textual training samples. Finally, the visual language model is fine-tuned on nuScenes data using these enriched pairs, enabling enhanced scene understanding and driving reasoning capabilities.}
    \label{fig:annotation}
\end{figure*}

\subsubsection{Teacher–Student Architecture}
Fig.~\ref{fig:annotation} details the data generation process utilizing the teacher OmniVLM, which serves as the foundational source for student model adaptation. The process begins with comprehensive knowledge mining, where ground-truth annotations, maneuvering signals, and domain-specific driving rules are systematically extracted from the Bench2Drive \cite{jia2024bench2drive} and nuScenes \cite{caesar2020nuscenes} datasets. For ground-truth annotations, dynamic obstacles are selected within approximately 15 meters in a 20-meter radius both ahead and behind the ego vehicle, as well as within approximately 30 meters in a 50-meter radius ahead and a 30-meter radius behind the ego vehicle, ensuring that the closest target in each lane is included. Traffic signs are annotated based on targets located within approximately 30 meters in a 30-meter radius ahead of the ego vehicle. Similarly, traffic lights are identified within approximately 30 meters in a 50-meter radius ahead of the ego vehicle. These elements jointly capture a diverse spectrum of environmental characteristics, including weather conditions, dynamic traffic participants, and static scene details, thereby enabling holistic modeling of complex real-world driving scenarios. Based on this structured knowledge base, the teacher OmniVLM is employed to automatically generate enriched textual descriptions, which incorporate environmental context, human-like attentional focus, and reasoning steps. This results in high-quality paired visual-textual data that supports downstream learning and model adaptation.

Subsequently, the fine-tuning stage centers on adapting a lightweight student OmniVLM with the curated data pairs from Bench2Drive and nuScenes. The student model’s streamlined design substantially reduces computational and memory overhead, facilitating deployment on resource-constrained platforms such as embedded automotive systems, without compromising its ability to comprehend scenes and reason about driving decisions. This teacher–student strategy not only augments interpretability and operational efficiency for autonomous driving tasks but also ensures rapid inference and adaptability in practical scenarios where hardware resources are limited.

A key design consideration within our architecture is the multi-view, multi-frame visual input strategy. Specifically, synchronized video streams from six cameras mounted on the ego vehicle are leveraged, providing comprehensive 360-degree coverage of the surrounding environment. Unlike conventional approaches that rely solely on front-view images and thus lack full situational awareness, our method employs spatiotemporally rich visual context to capture crucial peripheral and rear information for robust scene understanding. This multi-view, multi-frame paradigm enables the model to transcend the conditional independence assumptions of previous works and fully exploit the synergistic relationships between visual and language modalities.

To further advance human-like attentional semantics and dynamic environmental modeling, we propose a global multimodal alignment strategy. This method jointly considers multi-view, multi-frame visual features and fine-grained textual information, integrating them through a learnable similarity matrix. Rather than treating each camera view or frame separately, our alignment mechanism aggregates features across all views and temporal segments, aligning multi-view, multi-frame image embeddings with semantic text embeddings in a unified feature space. Adaptive weighting based on semantic relevance ensures construction of joint representations that are holistic and temporally aware. This strategy yields robust, distraction-resistant scene representations, which are essential for safe and effective autonomous driving.

\subsubsection{OmniScene Annotation}
Building upon the teacher–student architecture, our pipeline leverages the student OmniVLM’s reasoning capability to extract driver attentional information and generate semantic scene annotations from rich multi-view, multi-frame visual data. As depicted in Fig.~\ref{fig:architect}, the annotation extraction process can be formally described as:

\begin{equation}
    T = \mathcal{F}_{OmniVLM}(P, \{I_{i}^1, I_{i}^2, ...I_{i}^t\}),
\end{equation}
where $\mathcal{F}_{OmniVLM}(\cdot)$  represents the student OmniVLM, $P$ and $t$ represent the task-specific prompts and history steps, respectively. $\{I_{i}^1, I_{i}^2, ...I_{i}^t\}$ corresponds to the temporal visual streams from the ego vehicle’s $i$-view camera, with $i \in \{\text{front, front left, front right, back, back left, back right}\}$. $T$ is the generated textual scene description, which provides detailed, context-aware environmental information. 

% The objective of this process is to utilize targeted prompts and real-time multi-view video inputs to extract actionable and attentional information through the OmniVLM. This approach selectively emphasizes critical elements, such as pedestrians, traffic signals, and dynamic obstacles, while filtering out irrelevant background details to ensure that the outputs directly support informed driving decisions.

The proposed framework integrates targeted prompt conditioning with real‑time multi‑view video analysis in OmniVLM to selectively attend to semantically salient traffic agents, such as pedestrians, signal states, and moving obstacles, while suppressing background clutter, yielding scene representations optimized for driving decision support.

In our implementation, the fine-tuned student OmniVLM demonstrates strong capability in complex scene reasoning, generating precise and contextually relevant driving annotations. By interpreting visual scenarios under the guidance of task prompts, the model outputs textual descriptions that enrich the dataset with driver attentional cues. These annotations substantially enhance the interpretability of driving environments and improve the decision-making capabilities of downstream autonomous driving models.

\begin{figure}
    \centering
    \includegraphics[width=1.0\linewidth]{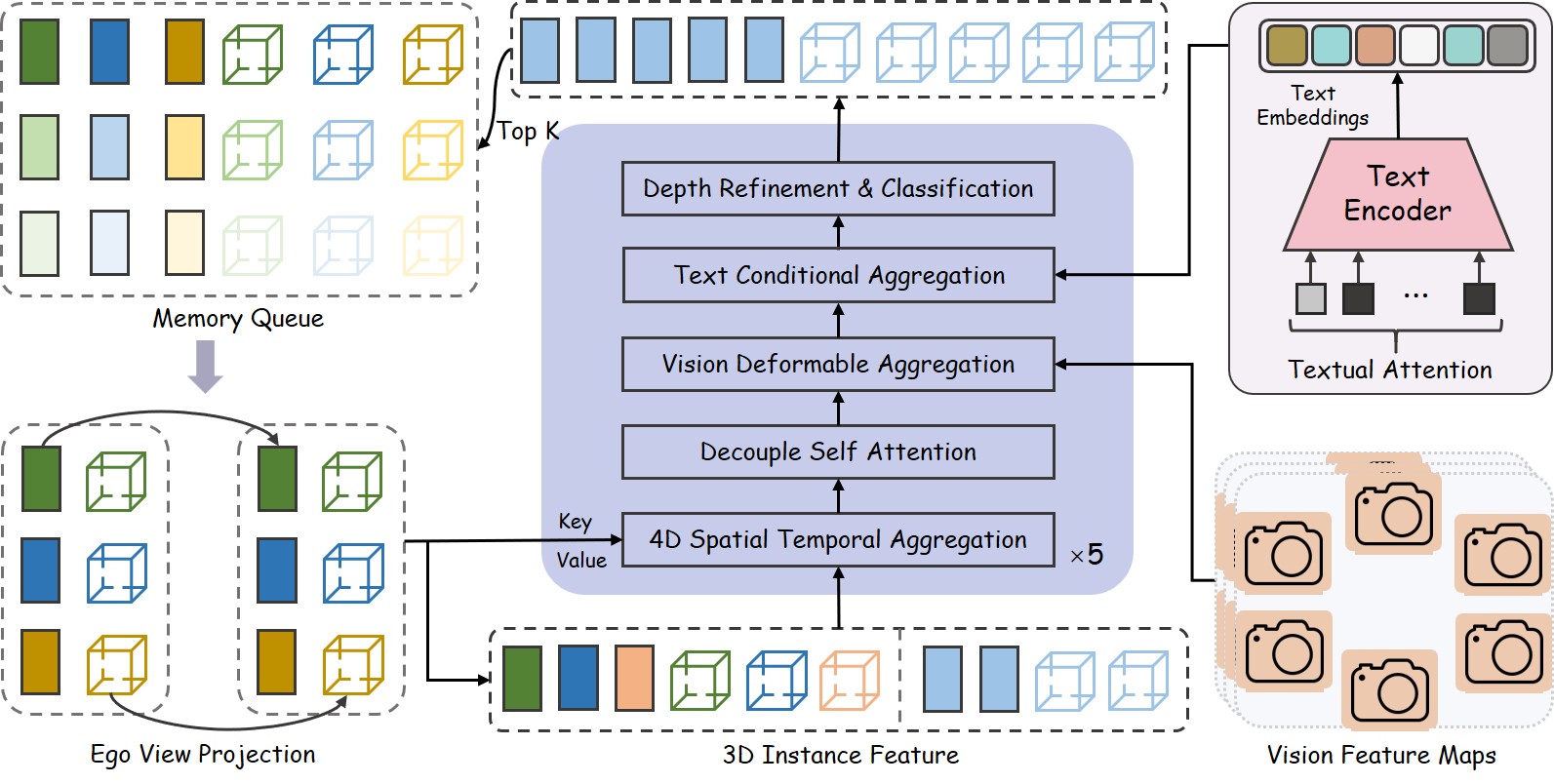}
    \caption{Hierarchical Fusion Strategy (HFS). This module incorporates both spatial-temporal fusion and text conditional aggregation to enable effective cross-modal integration.}
    \label{fig:fusion}
\end{figure}

\subsection{Hierarchical Fusion Strategy}
\label{sec:text-e2e}
Human-like attention encapsulates rich semantic cues from visual observations, offering complementary information to 3D instance features, which primarily encode geometric and structural properties. To achieve comprehensive scene understanding, we propose a hierarchical fusion strategy that effectively integrates these two modalities. The details of this fusion strategy are illustrated in Fig.~\ref{fig:fusion}.

\subsubsection{3D Instance Initialization}
\label{subsec:instance_initialization}

In this stage, we follow the sparse query-based paradigm proposed in SparseDrive~\cite{lin2022sparse4d, lin2023sparse4d, lin2023sparse4dv3}, aiming to efficiently initialize a set of 3D object instances in the scene using multi-view images.

% \paragraph{Learnable 3D Instance Queries}

We first initialize $N_{init}$ learnable 3D queries $\{\mathbf{q}_i\}_{i=1}^{N_{init}}$, where each query $\mathbf{q}_i \in \mathbb{R}^d$ encodes its spatial location $\mathbf{x}_i \in \mathbb{R}^3$, size $\mathbf{b}_i \in \mathbb{R}^3$, and semantic embedding. These queries are trainable and can adaptively shift to spatial regions of interest during training.

Given calibrated $M$ camera views and corresponding image features $\{\mathbf{I}_m\}_{m=1}^{M}$, each 3D query is projected into each view through camera projection $\mathrm{Proj}_m(\cdot)$. Multi-view features are sampled via bilinear interpolation:
\begin{equation}
\mathbf{f}_i^{(m)} = \mathrm{sample}(\mathbf{I}_m, \mathrm{Proj}_m(\mathbf{x}_i)), \quad m=1,2,\cdots,M.
\end{equation}
The sampled features are aggregated:
\begin{equation}
\mathbf{f}_i = \mathcal{A}\left( \{ \mathbf{f}_i^{(m)} \}_{m=1}^M \right),
\end{equation}
where $\mathcal{A}(\cdot)$ denotes the aggregation operation. For each query, we fuse the aggregated multi-view feature $\mathbf{f}_i$ and its query embedding to obtain the initial instance representation:
\begin{equation}
\mathbf{F}_i = [\mathbf{f}_i \,\|\, \mathbf{q}_i],
\end{equation}
where $[\cdot\,\|\,\cdot]$ denotes concatenation.

A proposal prediction head (PPH) is applied to $\mathbf{F}_i$ to predict object score $s_i$, 3D bounding box parameters $(\mathbf{x}_i, \mathbf{b}_i)$, and semantic label $c_i$:
\begin{equation}
(s_i,\, \mathbf{x}_i,\, \mathbf{b}_i,\, c_i) = \mathrm{PPH}(\mathbf{F}_i).
\end{equation}
Only proposals with $s_i > \tau$ are retained as valid 3D instances, where $\tau$ is a predefined threshold.

Compared to dense proposal generation, this sparse query-based initialization efficiently reduces computational complexity and enables the model to focus on informative regions in 3D space. The queries are learnable and can be optimized end-to-end, providing a strong basis for subsequent spatial-temporal reasoning and instance refinement.

\subsubsection{4D Spatial Temporal Aggregation}
\label{subsec:bev}

To robustly capture both temporal dynamics and spatial dependencies among multiple 3D instance features, {\model} employs a decoupled cross-attention mechanism over historical instance features, as well as a decoupled self-attention module. The input to this stage is a sequence of historical instance features:

\[
\left\{
\mathbf{F}_{t-T}^{(1)}, \mathbf{F}_{t-T+1}^{(2)},\cdots, \mathbf{F}_t^{(N)}
\right\},
\]
where $\mathbf{F}_{t'}^{(i)}$ denotes the feature embedding of the $i$-th 3D instance at time $t'$, and $N$ is the number of instances per frame.

We first apply decouple cross-attention to explicitly model temporal dependencies for each instance across multiple frames. For the $i$-th instance, the feature at the current step $\mathbf{F}_t^{(i)}$ attends to its own history $\left\{\mathbf{F}_{t-T}^{(i)},\mathbf{F}_{t-T+1}^{(i)}, \dots, \mathbf{F}_{t-1}^{(i)}\right\}$, thus capturing both long-term trends and recent dynamics. The temporal update is computed as:
\begin{equation}
\tilde{\mathbf{F}}_t^{(i)} = \sum_{k=0}^{T} \alpha_{ik} \mathbf{W}_V \mathbf{F}_{t-k}^{(i)},
\end{equation}
where
\begin{equation}
\alpha_{ik} = \mathrm{softmax}_k \left(
    \frac{
        (\mathbf{W}_Q \mathbf{F}_t^{(i)})
        \cdot
        (\mathbf{W}_K \mathbf{F}_{t-k}^{(i)})^\top
    }{
        \sqrt{d}
    }
\right),
\end{equation}
where $\mathbf{W}_Q$, $\mathbf{W}_K$, $\mathbf{W}_V \in \mathbb{R}^{d \times d}$ are learnable projection matrices, and $d$ is the feature dimension.

With the temporally updated features $\left\{\tilde{\mathbf{F}}_t^{(1)}, \tilde{\mathbf{F}}_t^{(2)},\dots, \tilde{\mathbf{F}}_t^{(N)}\right\}$, we further exploit spatial relationships within the current frame using decouple self-attention. This allows each instance to aggregate context from all other entities, modeling both local and global spatial interactions. Formally, for each instance:
\begin{equation}
\hat{\mathbf{F}}_t^{(i)} = \sum_{j=1}^{N} \beta_{ij} \mathbf{U}_V \tilde{\mathbf{F}}_t^{(j)},
\end{equation}
where
\begin{equation}
\beta_{ij} = \mathrm{softmax}_j \left(
    \frac{
        (\mathbf{U}_Q \tilde{\mathbf{F}}_t^{(i)})
        \cdot
        (\mathbf{U}_K \tilde{\mathbf{F}}_t^{(j)})^\top
    }{
        \sqrt{d}
    }
\right),
\end{equation}
with $\mathbf{U}_Q$, $\mathbf{U}_K$, $\mathbf{U}_V \in \mathbb{R}^{d \times d}$ being spatial attention parameters.

% \paragraph{Hierarchical Fusion and Reasoning.}

By stacking decouple cross attention for temporal modeling and decouple self attention for spatial aggregation, {\model} enables explicit, disentangled encoding of both temporal and spatial dependencies. This hierarchical design first incorporates long-term temporal information at the instance level, followed by detailed spatial contextualization within each frame. Such separation improves both the interpretability and expressiveness of scene modeling, which is critical for downstream tasks in autonomous driving.

\subsubsection{Vision Deformable Aggregation}
\label{subsec:vision_deformable}

To further enhance the representation of each 3D instance feature, we use a vision deformable aggregation module that adaptively aggregates informative cues from multi-view image features, guided by the geometric prior of each instance. Specifically, for each 3D instance $i$ at time $t$, we consider the temporally and spatially enhanced feature $\hat{\mathbf{F}}_t^{(i)}$ as well as multi-view vision features $\{\mathbf{I}_m\}_{m=1}^M$ from $M$ camera viewpoints.

For each instance, we project its 3D position into the image planes of all $M$ cameras, obtaining a set of 2D coordinates $\{(u_m^{(i)}, v_m^{(i)})\}_{m=1}^M$. Around each projected center, we predict a set of $K$ sampling offsets $\{\Delta \mathbf{p}_{m,k}^{(i)}\}_{k=1}^K$ based on $\hat{\mathbf{F}}_t^{(i)}$, so that the sampling locations are:
% \begin{equation}
% \mathbf{p}_{m,k}^{(i)} = (u_m^{(i)}, v_m^{(i)}) + \Delta \mathbf{p}_{m,k}^{(i)},\quad m=1,2,\cdots,M;\ k=1,2,\cdots,K.
% \end{equation}
\begin{equation}
\mathbf{p}_{m,k}^{(i)} = (u_m^{(i)}, v_m^{(i)}) + \Delta \mathbf{p}_{m,k}^{(i)},
\quad 
\begin{array}{l}
m = 1,2,\ldots,M,\\
k = 1,2,\ldots,K.
\end{array}
\end{equation}

At each sampled location $\mathbf{p}_{m,k}^{(i)}$ in camera $m$, we extract the corresponding image feature:
\begin{equation}
\mathbf{z}_{m,k}^{(i)} = \mathrm{bilinear\_interpolate}\left(\mathbf{I}_m, \mathbf{p}_{m,k}^{(i)}\right).
\end{equation}
All sampled features are fused using learned weights $\alpha_{m,k}^{(i)}$:
\begin{equation}
\mathbf{v}_t^{(i)} = \sum_{m=1}^{M} \sum_{k=1}^{K} \alpha_{m,k}^{(i)} \mathbf{W}_v \mathbf{z}_{m,k}^{(i)},
\end{equation}
where $\mathbf{W}_v \in \mathbb{R}^{d_v \times d}$ is a learnable projection, and weights $\alpha_{m,k}^{(i)}$ are normalized by softmax:
\begin{equation}
\alpha_{m,k}^{(i)} = \mathrm{softmax}_{m,k}\left( \mathbf{w}_\alpha^\top \, \mathbf{z}_{m,k}^{(i)} \right),
\end{equation}
with $\mathbf{w}_\alpha$ as a learnable vector.

The aggregated vision feature $\mathbf{v}_t^{(i)}$ is then fused with the 3D instance feature via a gating mechanism, yielding the final enhanced instance representation:
\begin{equation}
\mathbf{F}_t^{(i),\mathrm{final}} = \mathrm{concate}\left( \hat{\mathbf{F}}_t^{(i)},\, \mathbf{v}_t^{(i)} \right),
\end{equation}
where $\mathrm{concate}(\cdot)$ denotes the fusion operator.

This deformable aggregation module adaptively attends to the most informative spatial locations in multi-view images for each 3D instance, effectively leveraging both geometric cues and dense visual context. As a result, the final instance feature $\mathbf{F}_t^{(i),\mathrm{final}}$ incorporates rich visual semantics, which facilitates downstream tasks such as 3D detection and trajectory prediction.

\subsubsection{Text Conditional Aggregation}
\label{subsec:text_conditional}

To further enrich 3D instance representations with semantic context, we introduce a text conditional aggregation module. This module integrates textual semantic information into each 3D instance feature, enabling context-aware reasoning guided by textual cues. The inputs to this module are enhanced instance features $\{\mathbf{F}_t^{(i),\mathrm{final}}\}_{i=1}^N$ and text features $\mathbf{T} \in \mathbb{R}^{d_T}$, obtained from a pre-trained text encoder CLIP.

For each instance $i$, we first project both the vision-augmented feature $\mathbf{F}_t^{(i),\mathrm{final}}$ and the text feature $\mathbf{T}$ into a shared embedding space:
\begin{equation}
\mathbf{f}_i = \mathbf{W}_f \mathbf{F}_t^{(i),\mathrm{final}}, \quad \mathbf{t} = \mathbf{W}_t \mathbf{T},
\end{equation}
where $\mathbf{W}_f \in \mathbb{R}^{d' \times d}$ and $\mathbf{W}_t \in \mathbb{R}^{d' \times d_T}$ are learnable projection matrices, and $d'$ is the fusion feature dimension.

We model text-conditioned aggregation using a gated attention mechanism. The final text-enhanced instance feature is computed as:
\begin{equation}
\mathbf{F}_t^{(i),\mathrm{text}} = \mathbf{f}_i + \gamma_i \cdot \mathbf{t},
\end{equation}
where the gating coefficient $\gamma_i$ is adaptively generated based on both $\mathbf{f}_i$ and $\mathbf{t}$:
\begin{equation}
\gamma_i = \sigma \left( \mathbf{w}_\gamma^\top \left[ \mathbf{f}_i \,\|\, \mathbf{t} \right] + b_\gamma \right),
\end{equation}
where $\sigma(\cdot)$ denotes the sigmoid activation, $\mathbf{w}_\gamma$ and $b_\gamma$ are learnable parameters, and $\left[ \cdot \,\|\, \cdot \right]$ denotes vector concatenation.

By leveraging the guidance of text features, the proposed text conditional aggregation module enables the model to adaptively inject rich semantic knowledge into 3D instance representation. 

\subsubsection{Depth Refinement}
\label{subsec:depth_refinement}

To further enhance the geometric accuracy of 3D instance representations, we introduce a depth refinement module following the text conditional aggregation stage. This module aims to correct and refine the estimated depth for each instance by leveraging both the enriched instance features and auxiliary depth cues from multi-view images.

Given the text-enhanced instance feature $\mathbf{F}_t^{(i),\mathrm{text}}$ and the initial estimated depth $d_t^{(i),\mathrm{init}}$ of the $i$-th instance, we predict a residual depth correction through a lightweight regressor:
\begin{equation}
\Delta d_t^{(i)} = \mathrm{MLP}_{\mathrm{depth}}\left( \mathbf{F}_t^{(i),\mathrm{text}} \right),
\end{equation}
where $\mathrm{MLP}_{\mathrm{depth}}(\cdot)$ denotes a multi-layer perceptron specialized for depth adjustment.
The refined depth is then given by:
\begin{equation}
d_t^{(i),\mathrm{refined}} = d_t^{(i),\mathrm{init}} + \Delta d_t^{(i)}.
\end{equation}

% \paragraph{Multi-view Depth Consistency.}

To further regularize the instance depth, we enforce consistency with auxiliary depth maps $\{D_m\}_{m=1}^M$ predicted from the multi-view images. Specifically, for each instance, we project its refined 3D location onto the $m$-th camera's image plane to obtain the correspondence $(u_m^{(i)}, v_m^{(i)})$. The view-wise depth alignment loss is defined as:
\begin{equation}
\mathcal{L}_{\mathrm{depth}}^{(i)} = \frac{1}{M} \sum_{m=1}^{M} \left\| d_t^{(i),\mathrm{refined}} - D_m(u_m^{(i)}, v_m^{(i)}) \right\|_1.
\end{equation}
This loss encourages the refined instance depth to be consistent with the predicted scene geometry from all camera views.

The proposed depth refinement module effectively corrects geometric errors in 3D perception by adaptively regressing depth residuals and enforcing cross-view consistency. This design leverages the semantic, visual, and textual context aggregated in previous stages, further improving the reliability of downstream 3D understanding and prediction tasks.

\subsection{Vision-based End-to-End Model}
\label{sec:e2e}
\subsubsection{Multimodal Prediction and Planning}
{\model}’s multimodal trajectory prediction head integrates motion planning principles with learned trajectory forecasting to effectively capture the diverse future behaviors of agents in complex urban scenarios.

At inference time, the motion planning head operates over multiple future modes and temporal steps, generating a set of candidate future trajectories that include both ego and non-ego agents. The candidate motion modes are anchored via templates obtained from clustering datasets using KMeans \cite{steinley2006k}, stored as motion or plan anchors. These anchors encapsulate the prototypical maneuver patterns, such as straight, left-turn, right-turn, and yield, serving as structured priors that guide the generation and evaluation of predicted trajectories.

To enhance the expressivity of the model and disentangle interactions, {\model} optionally employs decoupled attention mechanisms, allowing for selective feature fusion between motion hypotheses and scene context. This facilitates robust reasoning in cases of ambiguous intent or occlusion.

The prediction module further leverages an instance queue for each tracked agent, maintaining a temporal buffer of feature embeddings across frames. The queue is parameterized by its capacity, embedding dimensionality, and tracking threshold, thus allowing the model to aggregate both local temporal dynamics and appearance information. Tracking enables the system to seamlessly update predictions as new sensory inputs are processed, reducing drift and improving long-horizon consistency.

At each planning cycle, future trajectories $\{\hat{\tau}_i^{(m)}\}_{m=1}^M$ are generated by matching current agent observations to the closest motion anchors and refining the candidate plans based on scene context, instance history, and the predicted likelihood of each maneuver. This approach supports diverse multimodal hypotheses, allowing the system to handle rare, complex, or long-tail behaviors common in real-world urban driving.

\subsubsection{Hierarchical Planning Selection}
Hierarchical planning selection is designed to identify the most feasible, safe, and contextually appropriate trajectory for both the ego agent and surrounding participants by holistically reasoning over multimodal predictions and scene constraints. After generating a set of diverse candidate trajectories $\{\hat{\tau}_i^{(m)}\}_{m=1}^M$ from the preceding prediction module, the planning head jointly reasons over these hypotheses, the evolving scene, and global navigation objectives. Each candidate trajectory is first projected onto a set of maneuver anchors $\{\mathcal{A}_k\}$, obtained via clustering methods such as KMeans, by measuring the similarity between the predicted path and each anchor using a distance metric:

\begin{equation}
    \text{sim}(\hat{\tau}_i^{(m)}, \mathcal{A}_k) = exp(-\frac{1}{T}\sum_{t=1}^{T}||\hat{y}_i^{t,(m)}-a_k^t||_2^2),
\end{equation}
where $\hat{y}_i^{t,(m)}$ denotes the predicted position at time $t$ for mode $m$, and $a_k^t$ is the corresponding anchor point.

Candidate trajectories that fail basic feasibility checks, such as violating drivable area constraints, intersecting with static obstacles, or conflicting with right-of-way rules, are masked out using an indicator function:
\begin{equation}
    f^{(m)} = \mathbb{I}[is\_feasible(\hat{\tau}_i^{(m)}, \mathcal{M}, \mathcal{O})],
\end{equation}
where $\mathcal{M}$ represents the map and $\mathcal{O}$ denotes surrounding objects.

For each remaining feasible trajectory, a composite utility score $s^{(m)}$ is computed by aggregating multiple criteria relevant to urban driving:
\begin{equation}
    s^{(m)} = \alpha_1 P(\hat{\tau}_i^{(m)}) - \alpha_2 J(\hat{\tau}_i^{(m)}) - \alpha_3R(\hat{\tau}_i^{(m)}) + \alpha_4,
\end{equation}
where $P$ measures advancement toward the planned route or target lane, 
$J$ quantifies motion comfort by penalizing sudden changes in acceleration or heading, $R$ evaluates expected collision or near-miss probability with other dynamic agents, and compliance rewards adherence to traffic rules and map constraints. The weights $\alpha_1$ to $\alpha_4$ are tuned to balance driving objectives, including safety, efficiency, and comfort.

The planning head is also informed by the temporal context maintained in the instance queue, which encodes historical intent and state transitions, thereby further refining score estimates and filtering spurious plans over time. As the system receives new observations, all candidate plans are continuously updated and rescored, allowing the planner to adapt to new obstacles and behavioral cues responsively.

Finally, the optimal trajectory is selected as:
\begin{equation}
    \hat{\tau}_i^{best} = \text{arg} \underset{m:f^{(m)}=1}{\text{max}} s^{(m)},
\end{equation}
with the plan being committed for near-term execution while maintaining closed-loop replanning at high frequency. This hierarchical, utility-driven framework enables {\model} to robustly handle complex, multi-agent urban scenarios and to anticipate both common and long-tail traffic events with interpretable, context-aware decision-making.

\subsubsection{Training Objectives}
To jointly optimize perception, motion prediction, and planning in an end-to-end manner, {\model} employs a unified loss function comprising several task-specific objectives. Each objective incorporates appropriate matching strategies and loss formulations to ensure effective multi-task learning.

We adopt the Hungarian algorithm to match each ground truth to one predicted detection. The perception loss, $\mathcal{L}_{det}$, is formulated as a weighted sum of a Focal loss for classification and an L1 loss for box regression:
\begin{equation}
    \mathcal{L}_{det} = \lambda_{det\_c}\mathcal{L}_{det\_c} + \lambda_{det\_r}\mathcal{L}_{det\_r},
\end{equation}
where $\mathcal{L}_{det\_c}$ denotes the detection classification loss, $\mathcal{L}_{det\_r}$ denotes the detection regression loss, and $\lambda_{det\_c}$, $\lambda_{det\_r}$ are their corresponding weights.

The map loss is defined similarly to the detection loss:
\begin{equation}
    \mathcal{L}_{map} = \lambda_{map\_c}\mathcal{L}_{map\_c} + \lambda_{map\_r}\mathcal{L}_{map\_r},
\end{equation}
where $\mathcal{L}_{map\_c}$ and $\mathcal{L}_{map\_r}$ are the classification and regression losses for mapping, and their weights are $\lambda_{map\_c}$ and $\lambda_{map\_r}$.

Depth regression adopts an L1 loss:

\begin{equation}
    \mathcal{L}_{depth} = \lambda_{depth}||d_{pred} - d_{gt}||_1,
\end{equation}
where $d_{pred}$ and $d_{gt}$ represent the predicted and ground truth depth, respectively.

Motion prediction minimizes the average displacement error (ADE) between multiple predicted trajectories and the ground truth, selecting the trajectory with the lowest ADE as the positive sample and treating the others as negatives. For planning, both the future ego status and intended path are predicted. Focal loss is used for classification, and L1 loss for regression:
\begin{align}
    \mathcal{L}_{motion} = \;&\lambda_{motion\_c}\mathcal{L}_{motion\_c} 
    \notag \\
    &+ \lambda_{motion\_r}\mathcal{L}_{motion\_r}, 
\end{align}
\begin{align}
    \mathcal{L}_{planning} = \;& \lambda_{plan\_c}\mathcal{L}_{plan\_c} \notag \\
    &+ \lambda_{plan\_r}\mathcal{L}_{plan\_r} \notag \\
    &+ \lambda_{plan\_{status}}\mathcal{L}_{plan\_{status}}, 
\end{align}
where the respective $\lambda$ parameters balance the classification, regression, and status prediction losses.

The overall loss for {\model} combines the above terms for multi-task training:
\begin{equation}
    \mathcal{L}_{total} = \mathcal{L}_{det} + \mathcal{L}_{map} + \mathcal{L}_{depth} +\mathcal{L}_{motion}+\mathcal{L}_{planning}.
    \label{eq:train}
\end{equation}

This multi-task objective encourages the model to learn effective representations for detection, mapping, depth estimation, and motion planning simultaneously, thereby improving autonomous driving planning ability.

\section{Experimental Settings}
\label{sec:experiments}
\subsection{Dataset}
% We evaluate our method on the nuScenes dataset \cite{caesar2020nuscenes}, a large-scale autonomous driving benchmark comprising 1,000 diverse driving scenes, each spanning 20 seconds with annotations provided at 2\,Hz. The dataset features a 360° multi-camera rig composed of six synchronized cameras (front, front-left, front-right, back, back-left, back-right) with minimal field-of-view overlap. Precise camera intrinsic and extrinsic parameters are provided for each frame to ensure accurate spatial alignment.

% The BEV occupancy labels $\{y_{t+1}, \cdots, y_{t+l}\}$ are generated by projecting 3D bounding boxes of dynamic agents onto the BEV plane, creating a spatiotemporal occupancy grid. All labels are transformed into the ego vehicle’s reference frame using GT future ego-motion, ensuring temporal consistency across frames.

We adopt the nuScenes benchmark \cite{caesar2020nuscenes}, a large‑scale multimodal dataset designed for autonomous driving research. It contains 1,000 diverse urban driving sequences, each lasting 20\, and densely annotated at 2\,Hz, covering a wide variety of traffic scenes, road layouts, and weather conditions. Data are captured with a full‑surround 360° sensor suite consisting of six synchronized cameras, a lidar, five radars, and an IMU/GNSS unit, providing complementary geometric and semantic cues. For the camera subsystem, per‑frame intrinsic and extrinsic calibrations are available, enabling accurate multi‑view spatial registration. The dataset includes 1.4\,M camera images, over 390\,k lidar sweeps, and fine‑grained 3D bounding box annotations for more than 23 object categories, including vehicles, pedestrians, bicycles, and traffic elements. Such richness, both in scale and sensor diversity, makes nuScenes a standard benchmark for evaluating perception, prediction, and planning algorithms in autonomous driving.

\begin{table*}
    \centering
    \caption{Perception (Detection) results on the nuScenes validation dataset.}
    \begin{tabular}{c|cccccccc}
    \toprule
    Method & Backbone  & mAP$\uparrow$ & NDS$\uparrow$ & mATE$\downarrow$ & mASE$\downarrow$& mAOE$\downarrow$& mAVE$\downarrow$& mAAE$\downarrow$\\
    \midrule
    UniAD \cite{hu2023planning}& ResNet101 & 0.380 &0.498 &0.684 &0.277& 0.383& 0.381 &0.192 \\
    SparseDrive \cite{sun2024sparsedrive}& ResNet50 & \textbf{0.418} &0.525 &0.566 & \textbf{0.275} &0.552 &0.261 &0.190\\
    \model & ResNet50 & \textbf{0.418} & \textbf{0.526} & \textbf{0.555} & 0.279 & \textbf{0.542} & \textbf{0.256} & \textbf{0.189}\\
    % \midrule
    % UniAD & ResNet101 & 0.380 &0.498 &0.684 &0.277& 0.383& 0.381 &0.192 \\
    % SparseDrive-B & ResNet101 & 0.496 &0.588 &0.543 &0.269& 0.376 &0.229& 0.179\\
    % \model & ResNet101 & \\
    \bottomrule
    \end{tabular}
    \label{tab:Perception}
\end{table*}

\begin{table}
    \centering
    \caption{Perception (Tracking) results on the nuScenes validation dataset.}
    \begin{tabular}{c|cccc}
    \toprule
    Method & AMOTA$\uparrow$  & AMOTP$\downarrow$ & Recall$\uparrow$ & IDS$\downarrow$\\
    \midrule
    ViP3D \cite{gu2023vip3d}& 0.217 & 1.625 & 0.363 & - \\
    QD3DT \cite{hu2022monocular}& 0.242 & 1.518 & 0.399 & - \\
    MUTR3D \cite{zhang2022mutr3d}& 0.294 & 1.498 & 0.427 & 3822 \\
    UniAD \cite{hu2023planning}& 0.359 & 1.320 & 0.467 & 906 \\
    SparseDrive \cite{sun2024sparsedrive}& \textbf{0.386} & 1.254 & 0.499 & 886 \\
    \midrule
    \model & 0.378 & \textbf{1.235} & \textbf{0.528} & \textbf{503} \\
    \bottomrule
    \end{tabular}
    \label{tab:Tracking}
\end{table}

\begin{table}
    \centering
    \caption{Prediction results on the nuScenes validation dataset.}
    \begin{tabular}{c|cccc}
    \toprule
    Method & minADE(m)$\downarrow$ & minFDE(m)$\downarrow$ & MR$\downarrow$ & EPA$\uparrow$\\
    \midrule
    Cons Pos. \cite{hu2023planning}& 5.80 &10.27 &0.347 &- \\
    Cons Vel. \cite{hu2023planning}& 2.13 &4.01 &0.318 &- \\
    Traditional \cite{gu2023vip3d}& 2.06& 3.02 &0.277& 0.209 \\
    PnPNet \cite{liang2020pnpnet}& 1.15& 1.95 &0.226 &0.222 \\
    ViP3D \cite{gu2023vip3d}& 2.05& 2.84& 0.246& 0.226 \\
    UniAD \cite{hu2023planning}& 0.71 &1.02 &0.151 &0.456 \\
    SparseDrive \cite{sun2024sparsedrive}& 0.62 &0.99 &0.136 &0.482 \\
    \midrule
    \model & \textbf{0.61} & \textbf{0.96} & \textbf{0.128} & \textbf{0.488} \\
    \bottomrule
    \end{tabular}
    \label{tab:Prediction}
\end{table}

\subsection{Metrics}
We conduct a comprehensive evaluation across multiple autonomous driving tasks following established benchmarks. 
3D object detection is quantified by mean Average Precision (mAP), the composite Detection Score (NDS), and error terms for translation (mATE), scale (mASE), orientation (mAOE), velocity (mAVE), and attribute prediction (mAAE). Multi‑object tracking is assessed using Average Multi‑Object Tracking Accuracy (AMOTA), Precision (AMOTP), recall, and Identity Switch Count (IDS). 
% For 3D object detection, we employ the standard nuScenes evaluation metrics including mean Average Precision (mAP) and the composite nuScenes Detection Score (NDS), along with error metrics for translation (mATE), scale (mASE), orientation (mAOE), velocity (mAVE), and attribute prediction (mAAE). The tracking performance is assessed using Average Multi-object Tracking Accuracy (AMOTA), Average Multi-object Tracking Precision (AMOTP), recall rate, and identity switches count. In online mapping evaluation, we compute Average Precision (AP) for three critical map elements - lane dividers, pedestrian crossings, and road boundaries - then derive the mean AP across all categories.

For motion prediction, our benchmark aligns with UniAD \cite{hu2023planning} and incorporates four key metrics: minimum Average Displacement Error (minADE), minimum Final Displacement Error (minFDE), Miss Rate (MR), and End-to-end Prediction Accuracy (EPA). The planning evaluation adopts two principal metrics: trajectory L2 error, which is consistent with VAD \cite{jiang2023vad} implementation, and collision rate. We identify and address two critical limitations in previous collision evaluation methodologies \cite{hu2023planning, jiang2023vad}. First, the conventional occupancy map approach with 0.5\,m grid resolution fails to accurately detect collisions with small obstacles due to quantization artifacts. Second, existing methods neglect the dynamic changes in the ego vehicle heading during motion. Our enhanced evaluation protocol overcomes these shortcomings by: (1) performing precise bounding box intersection tests between the ego vehicle and obstacles, eliminating grid quantization errors; and (2) incorporating yaw angle estimation from trajectory points to properly account for vehicle orientation changes. To ensure fair comparison, we re-evaluate baseline methods \cite{hu2023planning, jiang2023vad} using our improved collision detection framework with their official model checkpoints. This rigorous evaluation protocol provides a more accurate assessment of planning performance in complex driving scenarios.

% For visual question answering (VQA) evaluation, we adopt standard metrics: CIDEr (CI-r), BLEU-1 (BL-1), BLEU-4 (BL-4), METEOR (ME-R), ROUGE-L (RO-L), ensuring a comprehensive analysis of language understanding and multimodal alignment.

For VQA evaluation, performance is benchmarked using CIDEr (CI-r), BLEU-1 (BL-1), BLEU-4 (BL-4), METEOR (ME-R), and  ROUGE-L (RO-L), providing a multifaceted assessment of linguistic quality and vision–language alignment.

\subsection{Implementation Details}
% Our model utilizes a temporal context of 1.0 seconds of past information to predict the future trajectory over a 2.0-second horizon. In the nuScenes dataset, this corresponds to 3 frames of past context and 4 frames into the future, operating at a frequency of 2\,Hz. 
The model predicts future trajectories 2s ahead using 1s of historical context, which in nuScenes corresponds to 3 past and 4 future frames.
The teacher OmniVLM is developed by innovatively extending Qwen2.5VL 72B \cite{bai2025qwen2}, while the student OmniVLM is similarly constructed by enhancing Qwen2.5VL 7B. 

At each past timestep, the model receives 6 multi-view camera images, each with a resolution of $256 \times 704$ pixels. The perception backbone encodes these images and projects them into a unified sparse voxel space. We discretize the 100\,m $\times$ 100\,m $\times$ 6\,m scene centered around the ego vehicle into sparse pillars with a spatial resolution of 0.5\,m $\times$ 0.5\,m $\times$ 0.2\,m, resulting in an efficient sparse volumetric representation.

Training is conducted with the AdamW optimizer using a one-cycle learning rate schedule starting at $2.0 \times 10^{-4}$. We train the model for 10 epochs with a total batch size of 96, distributed over 8 Tesla A800 GPUs. Mixed precision training is applied to accelerate computation and reduce memory consumption.

\section{Results}\label{result}
\subsection{Quantitative Results}
\subsubsection{Perception}
Table~\ref{tab:Perception} and Table~\ref{tab:Tracking} present the perception results, including both detection and tracking performance, on the nuScenes validation set. Our proposed model attains the highest nuScenes detection score at 0.526 and achieves the lowest mean ATE of 0.555\,m, outperforming SparseDrive and UniAD in detection accuracy and localization precision. The model also delivers the lowest mAOE, mAVE, and mAAE, while maintaining competitive mAP and mASE scores, further validating its robust perception capabilities in complex urban environments.

For tracking, our model achieves the best scores in AMOTP, Recall, and identity switches, with an AMOTP of 1.235, a Recall of 0.528, and only 503 identity switches, significantly surpassing all prior baselines. Although SparseDrive reports a slightly higher AMOTA, our approach excels in tracking robustness by improving recall and reducing identity switches. These comprehensive results highlight the effectiveness of our model in delivering reliable detection and tracking performance for urban autonomous driving scenarios.

\subsubsection{Prediction}
Table~\ref{tab:Prediction} presents the prediction results on the nuScenes validation dataset. Our proposed method outperforms all existing baselines across all metrics. Specifically, our model achieves the lowest minADE and minFDE values of 0.61 and 0.96, respectively, indicating more accurate trajectory predictions. Furthermore, our approach attains the lowest miss rate of 0.128 and the highest EPA score of 0.488, demonstrating both superior reliability and enhanced efficiency in motion prediction. Notably, our method consistently surpasses previous state-of-the-art approaches, such as SparseDrive and UniAD, highlighting the effectiveness of our design in complex urban driving scenarios.

\subsubsection{Planning}
Table~\ref{tab:Planning} presents the planning performance on the nuScenes validation set, covering a variety of LiDAR-based, vision-based, and LLM-based methods. 
Our proposed method achieves the best results across almost all metrics. For L2 trajectory error, our model outperforms all competitors, reaching the lowest average value of 0.58\,m. Across all prediction horizons, the method achieves leading results with L2 errors of 0.28, 0.55, and 0.91\,m at 1, 2, and 3\,s, respectively, outperforming all existing approaches.
In terms of collision rate, our method maintains the lowest or near-lowest values across all time steps. The collision rate is 0\% at 1\,s and 0.04\% at 2\,s, indicating superior safety in short-term planning. At 3s, our model achieves a collision rate of 0.19\%, matching or surpassing other leading approaches.

Compared with recent strong vision-based methods such as GenAD, UAD, and SparseDrive, as well as LLM-based approaches like VLP-VAD and Senna, our method demonstrates clear advantages in both accuracy and safety. These results highlight the effectiveness and robustness of our approach for planning in challenging autonomous driving scenarios.
\begin{table*}
    \centering
    \caption{Planning results on the nuScenes validation dataset.}
    \begin{tabular}{c|l|cccc | cccc}
    \toprule
    ~ & \multirow{2}{*}{Method} & \multicolumn{4}{c|}{L2 (m)$\downarrow$} & \multicolumn{4}{c}{CR (\%)$\downarrow$} \\
    \cmidrule{3-10}
    ~ & ~ & 1s &2s &3s &Avg. &1s &2s &3s &Avg. \\
    \midrule
    \multirow{3}{*}{Lidar-based Method} & NMP \cite{zeng2019end} &0.53 &1.25 &2.67 &1.48 &0.04 &0.12 &0.87 &0.34 \\
    ~ & FF \cite{hu2021safe} & 0.55 &1.20& 2.54 &1.43 &0.06 &0.17 &1.07 &0.43 \\
    ~ & EO \cite{khurana2022differentiable}& 0.67 &1.36 &2.78 &1.60 &0.04 &0.09 &0.88 &0.33 \\
    \midrule
    \multirow{6}{*}{Vision-based Method} & ST-P3 \cite{hu2022st} & 1.33 &2.11 &2.90& 2.11& 0.23 &0.62 &1.27 &0.71 \\
    ~ & UniAD \cite{hu2023planning}& 0.45 &0.70 &1.04 &0.73 &0.62 &0.58 &0.63 &0.61 \\
    ~ & VAD \cite{jiang2023vad}& 0.41 &0.70 &1.05& 0.72 &0.03 &0.19 &0.43 &0.21 \\
    ~ & GenAD \cite{zheng2024genad} &0.28 &0.49 &0.78 &0.52 &0.08 &0.14& 0.34& 0.19 \\
    ~ & UAD \cite{guo2024end}& 0.39 &0.81 &1.50 &0.90& 0.01& 0.12 &0.43 &0.19 \\
    ~ & SparseDrive \cite{sun2024sparsedrive}&0.29 &0.58 &0.96 &0.61 &0.01& 0.05& \textbf{0.18} & \textbf{0.08} \\
    \midrule
    \multirow{8}{*}{LLM-based Method} & VLP-UniAD \cite{pan2024vlp}& 0.36 &0.68 &1.19 &0.74& 0.03 &0.12 &0.32 &0.16 \\
    ~ & VLP-VAD \cite{pan2024vlp} &0.30 &0.53 &0.84 &0.55 &0.01 &0.07 &0.38 &0.15 \\
    ~ & VLM-AD (UniAD) \cite{xu2024vlm} &0.39& 0.82 &1.43 &0.88 &0.05 &0.11 &0.43 &0.19 \\
    ~ & VLM-AD (VAD) \cite{xu2024vlm} &0.24 &0.46 &0.75 &0.48 &0.12 &0.17 &0.41 &0.23 \\
    ~ & Senna \cite{jiang2024senna} &0.37 &0.54 &0.86 &0.59 &0.09 &0.12 &0.33 &0.18 \\
    ~ & VAD $\&$ MiniCPM-V \cite{lu2025real}& 0.30& 0.48 &\textbf{0.67} & \textbf{0.48} &0.07 &0.10 &0.28 &0.15 \\
    ~ & VAD $\&$ Qwen-VL \cite{lu2025real}& 0.35 &0.53 &0.71& 0.53 &0.09 &0.12 &0.31 &0.17 \\
    \cmidrule{2-10}
    ~ & \model & \textbf{0.28} &\textbf{0.53} & 0.91 & 0.57 & \textbf{0.00} & \textbf{0.04} & 0.19 & \textbf{0.08}\\
    \bottomrule
    \end{tabular}
    \label{tab:Planning}
\end{table*}

\subsubsection{VQA Task}

Table~\ref{tab:qa} presents a comprehensive performance comparison on the nuScenes dataset across multiple benchmarks. Our models achieve substantial improvements over existing baselines on all evaluation metrics. The Ours 7B model achieves a CI-r score of 87.39, outperforming the best baseline InternVL3 14B at 70.01 by 24.9\%, and attains a BL-1 score of 38.4, which is 49.0\% higher than the best baseline Qwen2VL 72B at 25.76. The Ours 3B model achieves the highest BL-4 and RO-L scores of 7.42 and 28.97, with BL-4 showing a remarkable 66.5\% improvement over Qwen2VL 72B at 4.46, and RO-L increasing by 9.1\% over Qwen2VL 72B at 26.56. Across most tasks, both our 3B and 7B models consistently outperform mainstream models such as Qwen2.5VL and InternVL3, demonstrating the robustness and effectiveness of our approach. These results highlight the significant advancements our method offers for comprehensive scene understanding and reasoning in challenging urban environments.

\begin{figure*}
    \centering
    \includegraphics[width=0.85\linewidth]{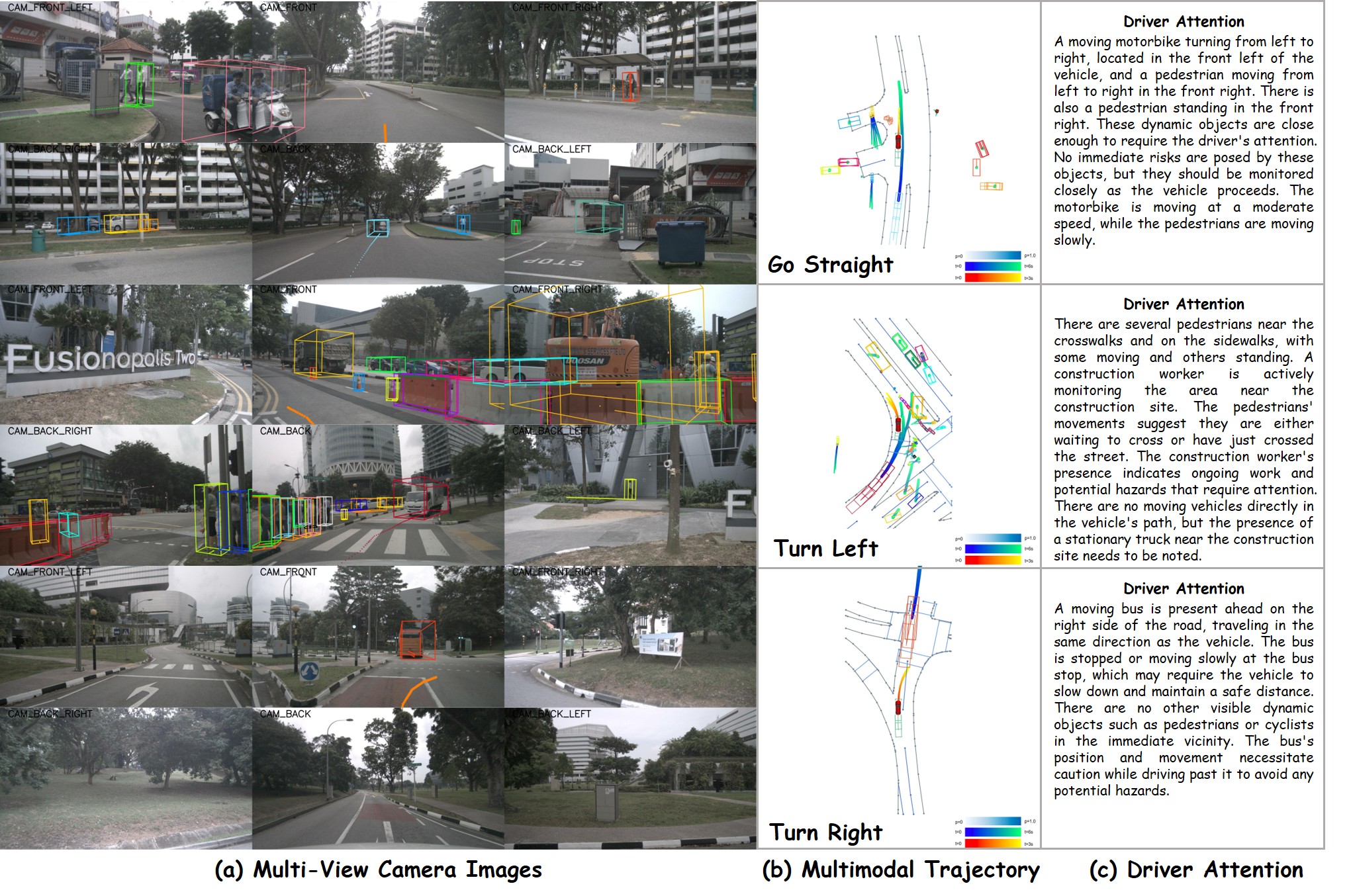}
    \caption{Visualization results under different driving intentions at intersections. {\model} learns different turning modes at intersections by jointly leveraging multi-view perception, trajectory predictions, and textual driver attention.}
    \label{fig1:planning}
\end{figure*}

\begin{table}
    \centering
     \caption{Comprehensive performance comparison on the nuScenes dataset. Our lightweight driving VLM outperforms prior works in all metrics.}
    \scalebox{1.0}{
    \begin{tabular}{c|ccccc}
    \toprule
        \textbf{Models} & \textbf{CI-r} & \textbf{BL-1} & \textbf{BL-4} & \textbf{ME-R} & \textbf{RO-L} \\
        \cmidrule{1-6}
        \multicolumn{1}{l|}{Qwen2.5VL 72B \cite{bai2025qwen2}} &67.14  &18.78  &3.25  &20.75  &21.91 \\
        \multicolumn{1}{l|}{Qwen2.5VL 32B} & 59.37 & 15.88& 1.72 & 17.69& 19.13\\
        \multicolumn{1}{l|}{Qwen2.5VL 7B} & 62.78& 19.86& 2.96& 22.52& 22.34\\
        \multicolumn{1}{l|}{Qwen2.5VL 3B} & 48.59  & 19.08 &1.91  &22.51 & 21.13\\
        \multicolumn{1}{l|}{Qwen2VL 72B \cite{team2024qwen2}} &  57.10 & 25.76&  4.46& 31.81& 26.56\\  
        \multicolumn{1}{l|}{InternVL3 14B \cite{zhu2025internvl3}}  & 70.01 & 8.82 & 1.09 & 74.15 & 19.18 \\
        \multicolumn{1}{l|}{InternVL3 8B } & 64.64 & 6.41 & 0.67 & 74.02 & 16.7\\

         \multicolumn{1}{l|}{LLaVA$\_$NEXT 7B \cite{li2024llava}} &53.54 & 4.71 & 0.47 & \textbf{76.09} & 15.63\\
        \cmidrule{1-6}
        \multicolumn{1}{l|}{OmniVLM 7B} & \textbf{87.39} & \textbf{38.4} & 6.88& \textbf{49.95}& 27.71\\
         \multicolumn{1}{l|}{OmniVLM 3B} & 60.26  & 28.3  & \textbf{7.42} & 40.53 & \textbf{28.97}\\
        \bottomrule
    \end{tabular}
    }
    \label{tab:qa}
\end{table}

% \begin{table}
%     \centering
%     \caption{Ablation study of text interaction guidance module on the nuScenes val dataset.}
%     \scalebox{0.9}{
%     \begin{tabular}{c|cccc | cccc}
%         \toprule
%         \multirow{2}{*}{Method} & \multicolumn{4}{c|}{L2 (m)$\downarrow$} & \multicolumn{4}{c}{CR (\%)$\downarrow$} \\
%         \cmidrule{2-9}
%         ~ &1s &2s &3s &Avg. &1s &2s &3s &Avg. \\
%         \midrule
%         w/o interaction & 0.30 & 0.58 & 0.95 & 0.61 & 0.02 & 0.05 & 0.22 & 0.10\\
%         w. interaction & \textbf{0.28} &\textbf{0.55} & \textbf{0.91} & \textbf{0.58} & \textbf{0.00} & \textbf{0.04} & \textbf{0.19} & \textbf{0.08}\\
%         \bottomrule
%     \end{tabular}
%     }
%     \label{tab:ab1}
% \end{table}

\begin{table*}
    \centering
    \caption{Ablation study of key designs in {\model} on the nuScenes validation set. “DR” denotes depth refinement; “EII” refers to ego instance initialization; “TDCA” corresponds to temporal decouple cross-attention; “SDCA” represents spatial decouple cross-attention; “TCA” indicates text conditional aggregation. Detection and tracking performance are reported under various model configurations, highlighting the contribution of each component to overall perception capability.}
    \scalebox{0.9}{
    \begin{tabular}{c|ccccc| ccccccc| ccc}
        \toprule
        \multirow{2}{*}{ID} & \multirow{2}{*}{DR} &\multirow{2}{*}{EII} &\multirow{2}{*}{TDCA} &\multirow{2}{*}{SDCA} &\multirow{2}{*}{TCA} & \multicolumn{7}{c|}{Perception (Detection)} & \multicolumn{3}{c}{Perception (Tracking)}\\
        \cmidrule{7-16}
        % \cline{8-17}
        ~ & ~ & ~ & ~ & ~ & ~ & mAP$\uparrow$ & NDS$\uparrow$& mATE$\downarrow$& mASE$\downarrow$& mAOE$\downarrow$& mAVE$\downarrow$& mAAE$\downarrow$& AMOTA$\uparrow$& AMOTP$\downarrow$ & Recall$\uparrow$\\
        \midrule
        1 &  & \checkmark & \checkmark & \checkmark &\checkmark & 0.415 & 0.521 & 0.577 & 0.278 & 0.552 & 0.271 & 0.192 & 0.379 & 1.266 & 0.486\\
        2 & \checkmark &  & \checkmark & \checkmark &\checkmark & 0.410 & 0.520 & 0.568 & 0.278 & 0.544 & 0.275 & 0.188 & 0.364 & 1.261 & 0.464\\
        3 & \checkmark & \checkmark &  &\checkmark& \checkmark & 0.310 & 0.369 &0.729 & 0.289 & 0.720 & 0.888 & 0.228 & 0.029 & 1.711 & 0.362\\
        4 & \checkmark & \checkmark & \checkmark && \checkmark & 0.315 & 0.372 & 0.724 & 0.287 & 0.714 & 0.900 & 0.229 & 0.036 & 1.709 & 0.362\\
        
        5 & \checkmark & \checkmark & \checkmark& \checkmark &  & 0.416& \textbf{0.526}& \textbf{0.554} & \textbf{0.276}& \textbf{0.529} & 0.265 & 0.193 & 0.376 & 1.254 & 0.512\\
        
        6 & \checkmark & \checkmark & \checkmark & \checkmark & \checkmark & \textbf{0.418} & \textbf{0.526} & 0.555 & 0.279 & 0.542 & \textbf{0.256} & \textbf{0.189} & \textbf{0.378} & \textbf{1.235} & \textbf{0.528} \\
        \bottomrule
    \end{tabular}
    }
    \label{tab:ab1}
\end{table*}

\begin{table*}
    \centering
    \caption{Ablation study of key designs in {\model} on the nuScenes validation set. Prediction and Planning performance are reported under various model configurations, highlighting the contribution of each component to overall prediction and planning capability.}
    \scalebox{0.95}{
    \begin{tabular}{c|ccccc| ccc |cccc | cccc}
        \toprule
        \multirow{2}{*}{ID} & \multirow{2}{*}{DR} &\multirow{2}{*}{EII} &\multirow{2}{*}{TDCA} &\multirow{2}{*}{SDCA} &\multirow{2}{*}{TCA} & \multicolumn{3}{c|}{Prediction} & \multicolumn{4}{c|}{Planning L2 (m)$\downarrow$} & \multicolumn{4}{c}{Planning CR (\%)$\downarrow$} \\
        \cmidrule{7-17}
        % \cline{8-15}
        ~ & ~ & ~ & ~ & ~ & ~ & minADE$\downarrow$& minFDE$\downarrow$ & MR$\downarrow$& 1s &2s &3s &Avg. &1s &2s &3s &Avg. \\
        \midrule
        1 & & \checkmark & \checkmark & \checkmark &\checkmark &0.62 & 0.98 & 0.13 & 0.29 & 0.57 & 0.94 & 0.60 & \textbf{0.00} & 0.06 & 0.22 & 0.09\\
        2 & \checkmark &  & \checkmark & \checkmark &\checkmark &0.62 & 0.97 & 0.14 & 0.29 & 0.57 & 0.93 & 0.60 & 0.01 & 0.08 & 0.24 & 0.11\\
        3 & \checkmark & \checkmark &  &\checkmark& \checkmark & 1.10 & 1.69 & 0.23 & 0.33 & 0.64 & 1.05 & 0.67 & 0.01 & 0.12 & 0.47 & 0.20\\
        4 & \checkmark & \checkmark & \checkmark && \checkmark & 1.11 & 1.71 & 2.33 & 0.32 & 0.62 & 1.03 & 0.66 & 0.03 & 0.12 & 0.41 & 0.19\\
        5 & \checkmark & \checkmark & \checkmark& \checkmark &  & 0.62&1.25& 0.51& 0.30 & 0.58 & 0.95 & 0.61 & 0.02 & 0.05 & 0.22 & 0.10\\
        6 & \checkmark & \checkmark & \checkmark & \checkmark & \checkmark & \textbf{0.61}& \textbf{0.96} &\textbf{0.13} &\textbf{0.28} &\textbf{0.55} & \textbf{0.91} & \textbf{0.58} & \textbf{0.00} & \textbf{0.04} & \textbf{0.19} & \textbf{0.08}\\
        \bottomrule
    \end{tabular}
    }
    \label{tab:ab1-1}
\end{table*}

\begin{table}[h]
    \centering
    \caption{Ablation study of the number of trajectory modes on the nuScenes validation dataset.}
    \scalebox{0.9}{
    \begin{tabular}{c|cccc|cccc}
    \toprule
         \multirow{2}{*}{Number of mode} &  \multicolumn{4}{c|}{L2 (m)$\downarrow$} & \multicolumn{4}{c}{CR (\%)$\downarrow$} \\
         \cmidrule{2-9}
        ~ &1s &2s &3s &Avg. &1s &2s &3s &Avg. \\
         \midrule
         1 &  \textbf{0.28} & 0.56 & 0.94 &0.59 & 0.01 & 0.08 & 0.28 & 0.12\\
         2 & 0.29 & 0.56 & 0.94 & 0.59 & 0.01 & 0.06 & 0.21 & 0.10\\
         3 & \textbf{0.28} & 0.56 & 0.92 &0.59 &0.04 & 0.07 & 0.18 & 0.10\\
         4 & \textbf{0.28} & 0.56 & 0.93 & 0.59 & 0.02 & 0.07 & 0.25 & 0.11\\
         5 & 0.29 & 0.57 & 0.95 & 0.60 & 0.01 & 0.06 & 0.23 & 0.10\\
         6 & \textbf{0.28} & \textbf{0.55} & \textbf{0.91} & \textbf{0.58} & \textbf{0.00} & \textbf{0.04} & \textbf{0.19} & \textbf{0.08} \\
         10 & 0.34 & 0.66 & 1.09 & 0.70 & 0.01 & \textbf{0.04} & 0.22 & 0.09\\
         \bottomrule
    \end{tabular}
    }
    \label{tab:ab2}
\end{table}

\begin{table*}
    \centering
    \caption{Quantitative results on the nuScenes validation dataset.}
    \begin{tabular}{c |cc|cccc|ccc|ccc}
    \hline
     & \multicolumn{2}{c|}{Perception} & \multicolumn{4}{c|}{Prediction} & \multicolumn{6}{c}{Planning} \\
    \cmidrule{2-13}
    Method & \multirow{2}{*}{IoU$_{Veh}$$\uparrow$} & \multirow{2}{*}{IoU$_{Ped}$$\uparrow$} & \multirow{2}{*}{IoU$\uparrow$} & \multirow{2}{*}{PQ$\uparrow$}& \multirow{2}{*}{SQ$\uparrow$} & \multirow{2}{*}{RQ$\uparrow$} & \multicolumn{3}{c|}{L2 (m)$\downarrow$} & \multicolumn{3}{c}{CR (\%)$\downarrow$}\\
    \cmidrule{8-13}
    ~  &~  &~  &~  &~ &~&~ & 1s & 2s & 3s & 1s & 2s & 3s \\ 
    \midrule
    ST-P3 \cite{hu2022st} &  38.79& 14.06 &36.89 & 29.10 & \textbf{69.77} & 41.71 & 1.33 & 2.11 & 2.90 & \textbf{0.23} & 0.62 & 1.27 \\
    \model & \textbf{39.08} & \textbf{17.49} & \textbf{38.54} & \textbf{29.83} & 69.56  & \textbf{42.88} & \textbf{1.22} & \textbf{1.94} & \textbf{2.68} & 0.26 & \textbf{0.60} & \textbf{1.17}\\
    \hline
    \end{tabular}
    \label{tab:generalization}
\end{table*}

\begin{table}[h]
    \centering
     \caption{Speed comparison of VLMs.}
    \begin{tabular}{c|cc}
    \toprule
         \multirow{2.6}{*}{Model} & \multicolumn{2}{c}{pixels = 256 $\times$ 256, tokens = 300} \\
         \cmidrule{2-3}
         ~ & Speed input (toks/s)& Speed output (toks/s)\\
         \midrule
         
         Qwen25VL 32B & 970.94	& 168.23 \\
         \midrule
         OmniVLM 7B	&2211.12& 294.31 \\
         OmniVLM 3B	&\textbf{3410.81}	&\textbf{391.43} \\
    \bottomrule
    \end{tabular}
    \label{tab:ab4}
\end{table}

\subsection{Qualitative Analysis}
Fig. \ref{fig1:planning} presents qualitative visualization results under different driving intentions at intersections. The proposed {\model} model jointly leverages multi-view perception, trajectory predictions, and textual driver attention to interpret complex intersection scenarios. The multi-view camera images capture diverse dynamic agents and static obstacles across various perspectives. The predicted multimodal trajectories indicate feasible future motions corresponding to different turning intentions, such as going straight, turning left, and turning right. Textual driver attention further provides detailed semantic interpretation by highlighting critical objects and contextual cues that influence the ego vehicle’s decision-making process, including pedestrians, construction workers, and parked buses. These comprehensive visualizations demonstrate that our approach can accurately perceive scene details, infer driving intentions, and provide interpretable reasoning for safe and reliable motion planning at challenging urban intersections.

\begin{figure*}
    \centering
    \includegraphics[width=0.85\linewidth]{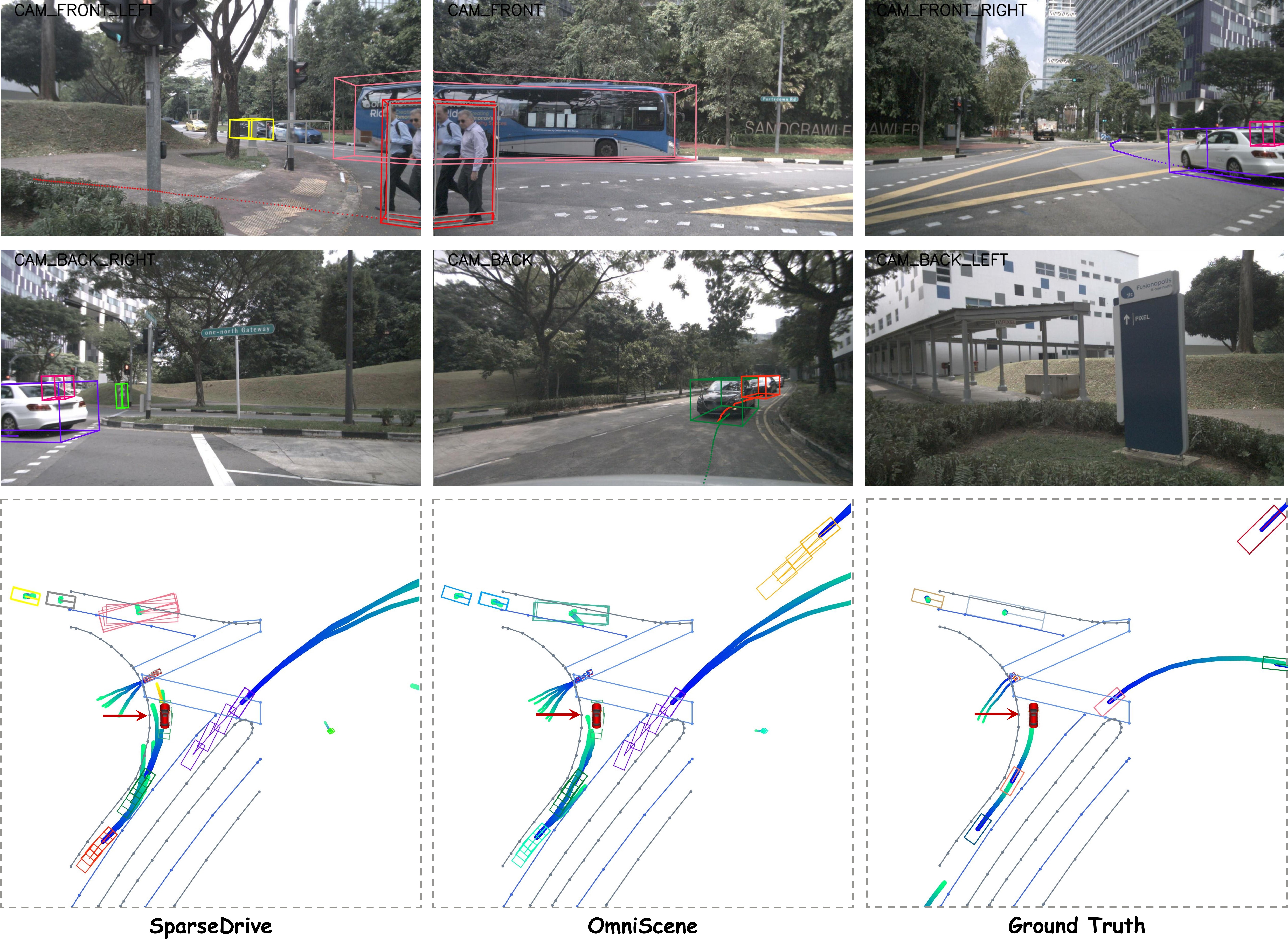}
    \caption{BEV visualization from SparseDrive, {\model}, and Ground Truth (left to right), depicting a scenario where multiple pedestrians appear in front of the ego vehicle, requiring emergency avoidance maneuvers.}
    \label{fig2:compare}
\end{figure*}

Fig. \ref{fig2:compare} illustrates qualitative BEV visualizations comparing SparseDrive, {\model}, and Ground Truth in a challenging scenario where multiple pedestrians appear in front of the ego vehicle, necessitating emergency avoidance maneuvers. The multi-view camera images capture the positions and movement of pedestrians and vehicles across the intersection. In the BEV maps, {\model} demonstrates improved trajectory prediction and obstacle localization, showing trajectories that closely match the ground truth and effectively adapt to the presence of dynamic agents. Compared to SparseDrive, {\model} provides more precise avoidance paths, indicating its enhanced ability to perceive critical obstacles and make safer, more reliable planning decisions. These results highlight the superiority of {\model} in handling complex urban scenarios with dense pedestrian activity and demanding safety requirements.

\begin{figure*}
    \centering
    \includegraphics[width=0.85\linewidth]{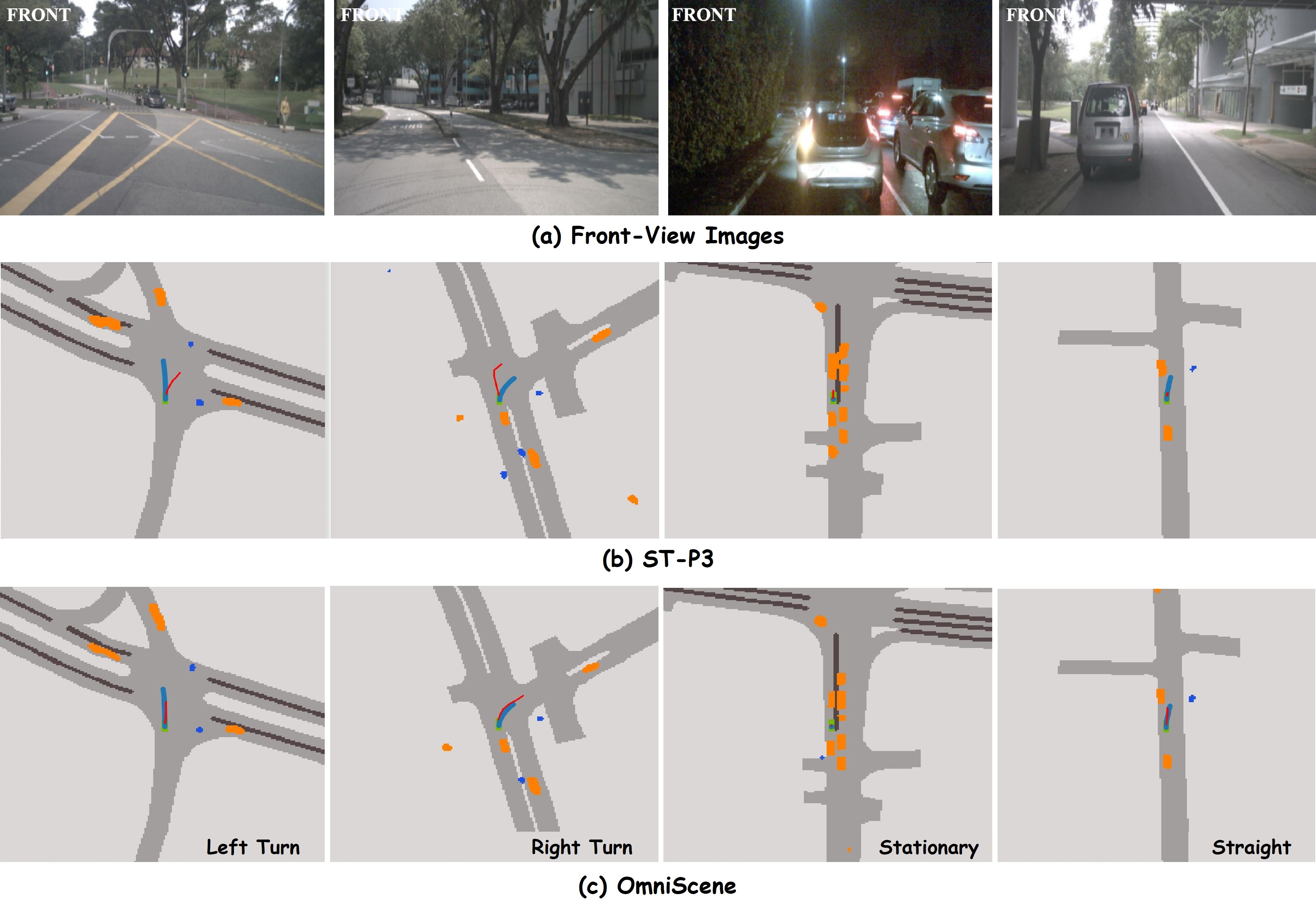}
    \caption{Qualitative comparison between ST-P3 and {\model}. The blue line indicates the GT trajectory, and the red line represents the predicted trajectory. The green object represents the ego vehicle, the yellow objects represent vehicles, and the dark blue objects represent pedestrians.}
    \label{fig:fig8}
\end{figure*}

\subsection{Ablation Study}
\subsubsection{Effectiveness of Designs in {\model}}
We conduct comprehensive ablation studies on the nuScenes validation set to evaluate the effectiveness of key architectural designs in {\model} across perception, prediction, and planning tasks. As reported in Table~\ref{tab:ab1}, each component, including depth refinement, ego instance initialization, temporal and spatial decouple cross-attention, and text conditional aggregation, contributes incrementally to overall detection and tracking performance. Notably, the temporal and spatial attention modules consistently enhance both mAP and NDS, resulting in improved object localization and classification. Likewise, incorporating text-conditioned aggregation yields substantial gains in tracking stability, as evidenced by improvements in AMOTA and Recall.

Table~\ref{tab:ab1-1} further demonstrates the impact of these design choices on prediction and planning outcomes. Integrating all proposed modules yields the lowest minADE and minFDE for trajectory prediction and the highest planning accuracy with reduced collision rates across all time horizons. In particular, the use of text conditional aggregation and cross-attention mechanisms enables the model to leverage contextual information more effectively, resulting in safer and more accurate motion planning.

Overall, the ablation results clearly show that the combination of multimodal textual cues, spatial and temporal attention modules, and refinement strategies in {\model} is crucial for achieving strong, balanced performance across the perception-planning pipeline in complex urban driving environments.

\subsubsection{Discussion on Multimodal Motion Planning}
Table~\ref{tab:ab2} presents an ablation study on the number of trajectory modes for multimodal motion planning. As the number of modes increases, both planning accuracy and safety are affected. The model achieves optimal performance when using six modes, with the lowest average L2 error of 0.58\,m and the lowest average collision rate of 0.08\%. Fewer trajectory modes, such as one or two, result in slightly higher L2 errors and collision rates, indicating limited motion diversity. Conversely, using ten modes leads to a notable increase in L2 error, reaching 0.70\,m, which suggests that introducing too many modes may introduce prediction uncertainty and reduce planning precision. These results demonstrate the importance of selecting an appropriate number of trajectory modes to strike a balance between prediction diversity and planning accuracy.

\subsubsection{Generalization on Other End-to-end Models}
To rigorously evaluate the generalization capability of our model, we integrate the Text Interaction Guidance Module into ST-P3 by fusing text features with BEV features, enabling the model to leverage both semantic and spatial information during perception and planning.

Table~\ref{tab:generalization} presents the quantitative results of this integration. We systematically evaluate perception, prediction, and planning, reporting IoU for vehicles and pedestrians in perception. For prediction, we provide results on Iou, Panoptic Quality (PQ), Segmentation Quality (SQ), and Recognition Quality (RQ). In planning, we assess both the L2 positional error at 1, 2, and 3\,s and the collision rate at the corresponding time horizons.

After introducing the text interaction guidance module, ST-P3 demonstrates notable improvements across most metrics. For perception, the integration leads to higher IoU for vehicles and pedestrians, and the overall IoU is also improved. In prediction, performance gains are observed in IoU, PQ, and RQ, while SQ remains stable.

In planning, the enhanced model achieves lower L2 errors across all time steps, reflecting more accurate trajectory prediction. Furthermore, the Collision Rate is reduced at 2 and 3\,s, indicating safer planning over longer horizons.

To further illustrate these improvements, we present four challenging scenarios from the nuScenes dataset, visualized in Fig.~\ref{fig:fig8}. In all four cases, ST-P3 produces incorrect future predictions, whereas our method generates accurate trajectories that closely adhere to the road layout. Specifically, in the first, third, and last scenarios, ST-P3 mistakenly suggests right turn, straight, and stationary intentions, respectively, while the SDV is actually performing a left turn, stopping, and moving straight. In contrast, our method accurately predicts the control actions for these cases, owing to the attention-based textual supervision provided by our module. These results highlight the limitations of relying solely on BEV features for future action prediction and demonstrate the complementary advantages introduced by incorporating textual guidance.

Collectively, both the quantitative and qualitative results validate the generalization capability and effectiveness of the proposed text interaction guidance module when applied to ST-P3, underscoring its potential to enhance both perception and planning in end-to-end autonomous driving models.

\subsubsection{Evaluation of Real-Time Performance}
The experimental setup involves testing OmniVLM 7B and 3B models on a single A800, while Qwen25VL 32B is assessed using two A800. As shown in Table \ref{tab:ab4}, OmniVLM 3B outperforms Qwen25VL 32B with substantial improvements in both input and output speeds, approximately a $3.51\times$ increase in input speed and a $2.33\times$ increase in output speed. These gains underscore OmniVLM 3B's efficiency in managing complex, multimodal inputs and generating outputs swiftly. On the A800 platform, OmniVLM 3B exhibits remarkable performance, processing 300 input tokens in just 88 ms and efficiently generating outputs when restricted to 10 to 20 tokens. Despite being tested on dual A800, Qwen25VL 32B falls behind OmniVLM 3B in computational efficiency. With the total processing time ranging from 113 ms to 139 ms, OmniVLM 3B aligns well with real-time requirements that are crucial for applications such as route optimization, obstacle detection, or collision avoidance. This model thus emerges as a plug-and-play module for real-time tasks in autonomous systems, offering a balanced combination of processing speed and communication efficiency.

\section{Conclusion}\label{conclusion}
% In this paper, we propose {\model}, a novel attention-augmented multimodal 3D Scene Understanding framework for autonomous driving. Our approach is motivated by the need to address key limitations in existing systems, such as modality imbalance in multi-sensor fusion, insufficient utilization of high-level semantic context, and the lack of interpretability in trajectory planning. To this end, we introduce a Vision-Text learnable weighted fusion strategy to dynamically balance geometric and semantic features, a spatiotemporal module to ensure temporal coherence in dynamic scenes, and a probabilistic future prediction module with attention-augmented trajectory refinement. These components collectively enable our framework to achieve robust and interpretable performance across perception, prediction, and planning tasks. Future work will explore a deeper integration based on OmniScene, achieving unified multimodal representation learning that seamlessly incorporates visual, textual, lidar, and radar data. Additionally, we plan to investigate strategies for improved generalization in long-tail and rare scenarios, leveraging the complementary strengths of different sensor modalities to enhance robustness and adaptability in complex real-world driving environments.

In this work, we introduced {\model}, an attention‑enhanced multimodal 4D scene understanding framework for end‑to‑end autonomous driving. By combining geometry‑aware 3D reasoning with high‑level 4D semantic abstraction from vision–language modeling, and aligning them through hierarchical fusion and human‑like attentional mechanisms, {\model} produces unified, interpretable scene representations that improve perception, prediction, and planning in complex driving environments. Leveraging a teacher–student OmniVLM design, the framework efficiently transfers fine‑grained attentional knowledge to lightweight models, enabling deployment without compromising performance. Extensive experiments on the nuScenes benchmark demonstrate clear performance gains over state‑of‑the‑art baselines, highlighting the effectiveness of human‑aligned multimodal fusion in safety‑critical reasoning. Future work will explore broader multimodal integration within the OmniScene paradigm and investigate strategies to enhance generalization under long‑tail distributions and rare traffic scenarios.

% \end{thebibliography}
\bibliographystyle{IEEEtran}
\bibliography{main}

% \clearpage

% Biography text here without a photo.

\end{document}